\documentclass{article} %
\usepackage{colm2024_conference}

\usepackage{microtype}
\usepackage{hyperref}
\usepackage{url}
\usepackage{graphicx}
\usepackage{booktabs}
\usepackage{subcaption}
\usepackage{arydshln}
\usepackage{xspace}
\usepackage{amsmath}
\usepackage{amssymb}
\usepackage[capitalize]{cleveref}
\usepackage{makecell}
\usepackage{pifont}
\usepackage{wrapfig}

\definecolor{darkblue}{rgb}{0, 0, 0.5}
\hypersetup{colorlinks=true, citecolor=darkblue, linkcolor=darkblue, urlcolor=darkblue}

\title{Inspecting and Editing Knowledge Representations in \\ Language Models}

\author{Evan Hernandez, Belinda Z. Li \& Jacob Andreas \\
Computer Science \& Artificial Intelligence Laboratory \\
Massachusetts Institute of Technology\\
\texttt{\{dez,bzl,jda\}@mit.edu} \\
}

\newcommand{\ourmethod}{\textsc{remedi}\xspace}
\newcommand{\feat}{f}
\newcommand{\phum}{p_{\text{H}}}
\newcommand{\plm}{p_{\text{LM}}}
\newcommand{\corr}{r(\plm, \phum)}

\colmfinalcopy %
\begin{document}

\maketitle

\begin{abstract}
Neural language models (LMs) represent facts about the world described by text.
Sometimes these facts derive from training data (in most LMs, a representation of the word \emph{banana} encodes the fact that bananas are fruits). Sometimes facts derive from input text itself (a representation of the sentence \emph{I poured out the bottle} encodes the fact that the bottle became empty).
We describe \ourmethod, a method for learning to map statements in natural language to fact encodings in an LM's internal representation system.
\ourmethod encodings can be used as \emph{knowledge editors}: when added to LM hidden representations, they modify downstream generation to be consistent with new facts. \ourmethod encodings may also be used as \emph{probes}: when compared to LM representations, they reveal which properties LMs already attribute to mentioned entities, in some cases making it possible to predict when LMs will generate outputs that conflict with background knowledge or input text.
\ourmethod thus links work on probing, prompting, and LM editing, and offers steps toward general tools for fine-grained inspection and control of knowledge in LMs.
\end{abstract}

\section{Introduction}

\begin{wrapfigure}{r}{.45\textwidth}
    \vspace{-4.5em}
    \includegraphics[width=.45\textwidth, trim=.2in 1.7in 19in 6in,clip]{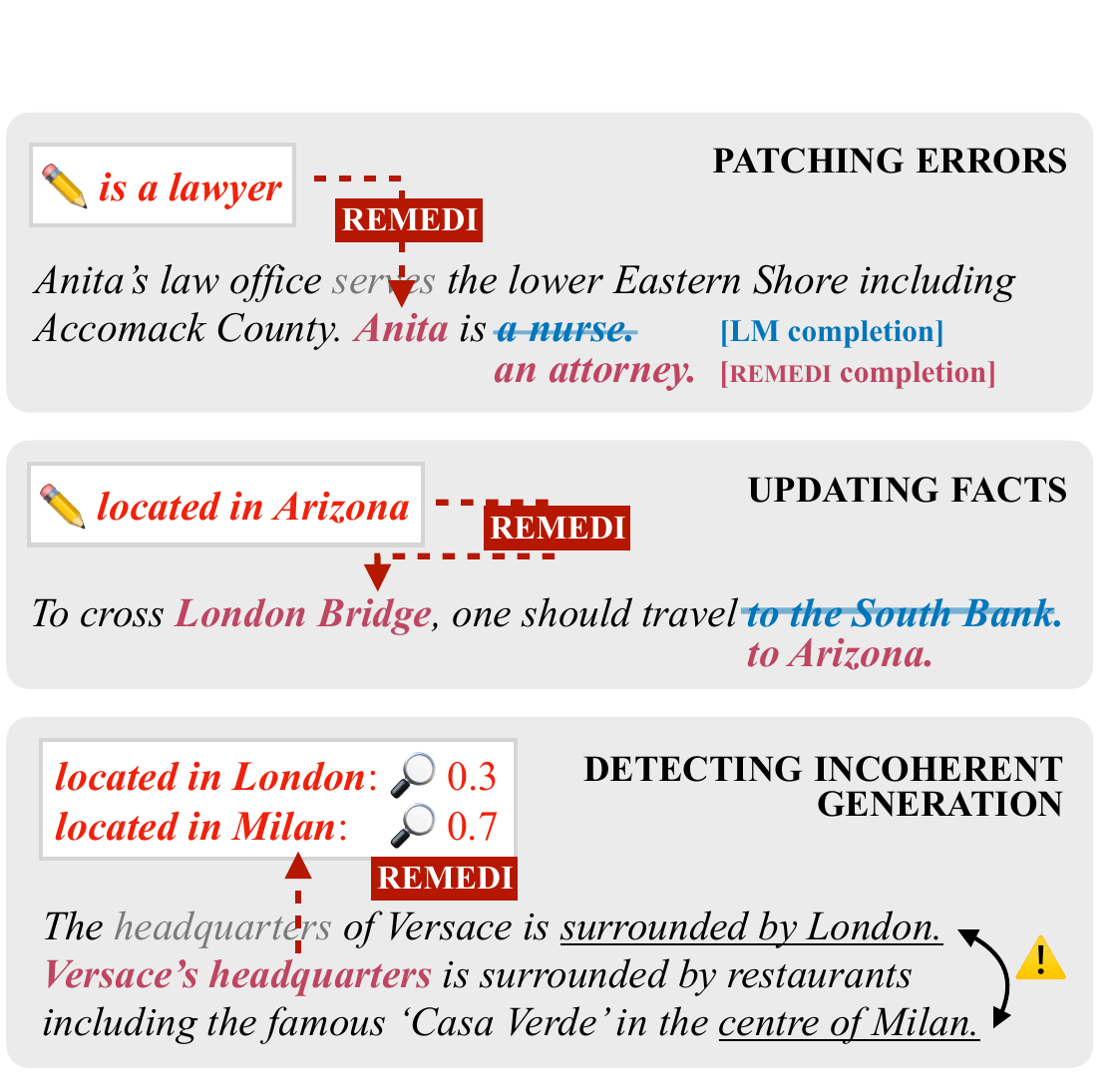}
    \caption{
    \ourmethod can patch errors made by an LM and insert new facts with or
    without context provided in the prompt. It can also help detect errors
    before generation.
    }
    \label{fig:teaser}
    \vspace{-1em}
\end{wrapfigure}

Neural language models (LMs) build implicit, structured models of the state of
the world: their representations encode general knowledge \citep{petroni-etal-2019-language} and situations described in input text
\citep{li-etal-2021-implicit}.  Sometimes these representations contain
mistakes, resulting in errors in generated
text (\cref{fig:teaser}).  As LMs improve, versions of these problems are likely to
persist: large LM training sets contain erroneous and contradictory information,
go out of date, and harbor unexpected biases \citep{parrots}. Even in domains
where LM generation is more reliable, understanding how model-internal
representations relate to output is crucial for attribution and controlled
generation \citep{akyurek_fact_tracing,dai-etal-2022-knowledge}.  There is thus
a fundamental need for techniques that can
inspect and edit LMs' knowledge, whether derived from training data or input text.

This paper introduces \ourmethod (\textsc{re}presentation \textsc{medi}ation), a
technique for discovering directions in LM-internal representation spaces corresponding to encodings of
factual attributes (like \emph{is a lawyer} in \cref{fig:teaser}). When these 
encodings are added to LMs'
representations of entities (like \emph{Anita}), they \emph{edit} the facts that LMs attribute to
those entities---in some cases producing output that cannot be
produced with a corresponding textual prompt.  Encodings produced by \ourmethod
can also be used to \emph{interpret} LM representations, 
making it possible to probe LMs' factual knowledge, and to predict when they
will generate incorrect or incoherent output.

Even when trained only to modify LMs' background knowledge, \ourmethod generalizes to tasks that require querying and modifying knowledge specified in-context.
Our findings thus suggest that LMs represent and integrate information from these sources in a unified manner.
\ourmethod offers steps towards tools that can monitor and control language generation by %
directly specifying facts and situations in an LM's native encoding scheme.%
\footnote{Code and data are available at \url{https://github.com/evandez/REMEDI}.}

\section{\ourmethod}
\label{sec:method}

\newcommand{\lm}{p_{\text{LM}}}
\newcommand{\h}{h}
\newcommand{\layer}{\ell}
\newcommand{\str}{x}
\newcommand{\entity}{\text{entity}}
\newcommand{\attr}{\text{attr}}
\newcommand{\direction}{d}
\newcommand{\editor}{F}
\newcommand{\loss}{\mathcal{L}}
\newcommand{\target}{t}
\newcommand{\hedited}{z}
\newcommand{\mediated}{\text{tgt}}
\newcommand{\prior}{\text{prior}}

\newcommand{\entitystr}{\str_{\entity}}
\newcommand{\attrstr}{\str_{\attr}}
\newcommand{\lmlayer}{\lm^{(\layer)}}
\newcommand{\hlayer}{\h^{(\layer)}}
\newcommand{\hent}{\h_{\entity}}
\newcommand{\hentlayer}{\hlayer_{\entity}}
\newcommand{\hattr}{h_{\attr}}
\newcommand{\hattrlayer}{\hlayer_{\attr}}
\newcommand{\dattrlayer}{\direction^{(\layer)}_{\attr}}
\newcommand{\targetmed}{\target_{\mediated}}
\newcommand{\targetprior}{\target_{\prior}}
\newcommand{\lossmed}{\loss_{\mediated}}
\newcommand{\lossprior}{\loss_{\prior}}
\newcommand{\losskl}{\loss_{\text{KL}}}
\newcommand{\lamprior}{\lambda_1}
\newcommand{\lamkl}{\lambda_2}

\paragraph{Motivations: control and interpretability}
Consider the examples from \cref{fig:teaser} (top). In the first example, the LM is \textbf{prompted} with the text \textit{Anita's \underline{law office} serves the lower Eastern Shore\dots}, which provides some \textbf{context} about the \textbf{entity} Anita. However, when the LM generates a continuation of this prompt, it asserts that Anita is a nurse.
We term this incoherence a failure of \textbf{context integration}: information provided in the textual context has failed to alter the LM's predictions. It would be useful to identify and fix such errors, changing a model's encoding of entities like \textit{Anita} to ensure that she is correctly described as an \textit{attorney}.
In addition to ensuring discourse coherence,
it is often desirable to modify prior associations in LMs. In \cref{fig:teaser}
(middle) the LM strongly associates  \textit{London Bridge} with the city of
\textit{London} because the most famous London Bridge is located there. However,
there could be (and are\footnote{Such as the one in Lake Havasu City, Arizona.})
other London Bridges, and we might wish to control an LM to make the lesser-known bridge more salient.

It is sometimes possible to achieve these goals by carefully prompting models with the right input text.
But due to the non-systematic, opaque nature of prompt engineering \citep{jiang-etal-2020-know}, significant manual effort is often required to find a prompt (if one exists at all) that yields correct behavior and generalizes to new use cases.%
\footnote{These issues are not solved with scale: ``prompt injection attacks'' that cause LMs to ignore initial instructions \citep{ignore_previous_prompt,more_than_you} may also be viewed as failures of context integration, and might (beyond the scope of this paper) also be mitigated with better tools for directly manipulating representations of tasks rather than facts.}
Techniques for localizing generation failures within LMs' internal representations would make it possible to detect them in advance, and guide research aimed at mechanistic understanding of the relationship between LMs' internal representations and their textual outputs.

\begin{wrapfigure}{r}{.45\textwidth}
    \vspace{-2.75em}
    \includegraphics[width=.45\textwidth, trim=.2in 9in 19in .2in,clip]{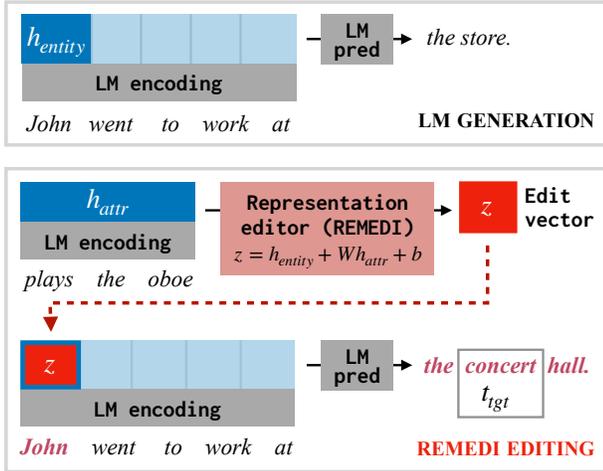}
    \caption{
    Illustration of \ourmethod. Given an prompt (\textit{John went to work at})
    and a desired attribute (\textit{plays the oboe}), \ourmethod constructs an
    edit to the hidden representation of \textit{John} that increases the
    probability of an appropriate completion (\emph{the concert hall}).
    }
    \label{fig:method}
    \vspace{-1em}
\end{wrapfigure}

\paragraph{Overview}
At a high level, our proposed approach learns how to intervene in an LM's
representation space to modify the LM's knowledge about a mentioned entity
(like \emph{Anita} in \cref{fig:method}). This intervention
ultimately updates the LM's representation of the entity to encode an
\textbf{attribute} (e.g., \textit{is a lawyer}) so that the LM will generate text about the entity consistent with the new attribute. This update operation can be specified by a single vector, and is applied to the hidden representation of a single token at a single layer. 
Edits produced by \ourmethod can also be applied out-of-context
(enabling controlled generation without textual prompts). By comparing edit vectors to unedited representations, \ourmethod additionally makes it possible to inspect representations of entities and attributes produced during ordinary model
operation.

\paragraph{Editing representations}
Assume we have a language model $\lm(\str)$ that assigns probabilities to
strings $\str$ consisting of tokens $\str_{1:n}$. In this paper, $\lm$ will
always be an autoregressive transformer \citep{vaswani-transformer} pretrained
on English text, as in the GPT family of models \citep{radford2019language, brown-gpt3}. These models decompose $p(\str)$ into a product of next-token distributions given preceding context: $\lm(\str) = \prod_i \lm(\str_i \mid \str_{1:i-1})$. Our goal is to modify an LM's internal state to cause it to generate desired text about a target entity. 

Where and how should we perform this modification? LMs encode factual information in hidden representations of entity mentions:
entities' states \citep{li-etal-2021-implicit}, perceptual
features \citep{abdou-etal-2021-language}, and other semantic properties
\citep{grand-semantic} have been shown to be \emph{linearly decodable} from
entity representations. 
To ensure that an LM encodes the fact \emph{Anita is a lawyer}, it should thus suffice to find an appropriate transformation of the representation of the token \emph{Anita}.

Formally, we denote the transformer's hidden representation for token $x_i$ in
layer $\ell$ as $\smash{\hlayer_i}$, and we write $\smash{\lm(\str \mid \hlayer_i = z)}$ to
mean the output of $\lm$ with $\hlayer_i$ ``hard-coded'' to equal $z$ during the
forward pass.\footnote{Henceforth we will omit the layer index, but the reader should always assume all operations occur at a single layer.}
Given representations of the entity $\hent$ and the target attribute $\hattr$,
\ourmethod specifically learns an \textbf{affine transformation} $\editor$ that returns a new entity representation $\hedited$ according to:
\begin{equation}
\label{eq:editor}
\hedited = \editor(\hent, \hattr) = \hent + W\hattr + b ~ .
\end{equation}
such that when $\hedited$ replaces the entity inside the LM, the LM will generate text consistent with the target.

How should we pick $\hattr$, $W$ and $b$?
Building on the success of linear probing approaches \citep{conneau-etal-2018-cram, belinkov-glass-2019-analysis}, it is tempting to begin by training a \emph{classifier} for the presence or absence of attributes.
For example, following \citet{li-etal-2021-implicit}, we could 
take $\hattr$ to be the LM's own representation of an attribute (like \emph{plays the oboe}; \cref{fig:method}), then
optimize $W$ and $b$ to predict whether an entity representation encodes the attribute:
\begin{equation}
\label{eq:probe}
p(\textrm{attribute} \mid \textrm{entity}) = \sigma(\hent^\top W \hattr + b) ~ .
\end{equation}
However, even when an LM encodes information in its representations, this information may not \emph{causally influence} subsequent generation \citep{ravfogel-etal-2020-null,elazar-etal-2021-amnesic, ravichander-etal-2021-probing}. An effective editor must identify fact encodings that are causally linked to output.

\textbf{Learning Effective Edits}\hspace{1em}
\ourmethod optimizes $W$ and $b$ to directly intervene in an LM. 
We assume access to a dataset of tuples $(\str_{1:n-1}, i_\text{entity}, t_\text{attr}, \targetmed)$, where $\str_{1:n-1}$ is a textual \textbf{context}
 (e.g.\ \emph{John went to work at}), $i_\text{entity}$ is the index of an
 entity within the context, $t_\text{attr}$ is the attribute to be
 inserted (\emph{plays the oboe}), and $\targetmed$ is a \textbf{generation
 target}: a completion that should be assigned high probability if the attribute
 is applied to $x_\text{entity}$ (\emph{the concert hall}).
Following \citet{li-etal-2021-implicit}, we obtain a representation
$\hattr$ by averaging the LM's encoding of $t_\text{attr}$.
We train the editor $\editor$ to maximize the probability that $\lm$ assigns to $\targetmed$ after modifying the hidden representation of $\str_\text{entity}$(\cref{fig:method}):
\begin{equation}
    \lossmed(\hedited) = -\log \lm(\str_n = \targetmed \mid \str_{1:n-1}, \hent =
    \hedited) ~ .
\end{equation}
\textbf{Learning Non-Destructive Edits}\hspace{1em}
When LMs encode strong prior associations between entities and properties (e.g., in the \textit{London Bridge} example; see \citealp{hase-beliefs}), it is necessary to remove these facts while inserting new ones.
We obtain a target string $\targetprior$ assigned a high
pre-edit probability, and train $\editor$ to minimize the probability of $\targetprior$:
\begin{equation}
    \lossprior(\hedited) = \log \lm(\str_n = \targetprior \mid \str_{1:n-1}, \hent =
    \hedited) ~ .
\end{equation}
Finally, to prevent the degenerate solution in which the language model always (and only) predicts $\targetmed$, we penalize the language model for changing its distributions on all tokens between the entity mention and the time at which it predicts $\targetmed$:
\begin{equation}
    \losskl(\hedited) = \sum^{\str_i \neq \entitystr}_{\str_i} D_{\text{KL}}\Big(
        \lm(\cdot \mid \str_{<i},\  \hent = \hedited) \ \Big\| \
        \lm(\cdot \mid \str_{<i})
    \Big) ~ .
\end{equation}
Unlike the distribution over tokens at the end of the prompt, which should
change dramatically under the intervention, the distribution over these
intermediate tokens should not change significantly. $\losskl$ penalizes such changes.
Thus, the complete objective function is:
\begin{equation}
    \label{eq:objective}
    \loss(\hedited) = \lossmed(\hedited) + \lamprior\lossprior(\hedited) +
    \lamkl\losskl(\hedited) ~ ,
\end{equation}
where $\lamprior$ and $\lamkl$ are hyper-parameters.
We evaluate \ourmethod by studying its ability to control model output (\cref{sec:generation}) and to interpret and predict model behavior (\cref{sec:probing}).

\section{Related Work}
\label{sec:related}

\paragraph{Probing factual knowledge}
Large language models (LLMs) trained on massive text datasets have been shown to encode context-agnostic factual knowledge, which can be queried through a text prompt~\citep{petroni-etal-2019-language}.
Most work on extracting background factual knowledge from LMs focuses on designing textual \textit{queries} for different sources of knowledge~\citep{richardson-sabharwal-2020-qa,peng2022copen}.
Additionally, \emph{knowledge probes} may sometimes recover factual information even in cases when LMs do not generate truthful outputs with high probability
\citep{burns2022dl}.

\paragraph{Probing representations of individual situations}
Neural LMs have also been shown to build representations of context-dependent knowledge.~\citep{li-etal-2021-implicit} show that they track aspects of entity state over a discourse, and this state can be extracted from LM representations of contextualized entity tokens. 
Furthermore, many LMs have been (indirectly) evaluated on their ability to track context-dependent knowledge by having their performance measured on downstream \textit{reading comprehension} tasks in wich the LM is expected to answer questions about facts within a discourse. Reading comprehension datasets such as CoQA~\citep{reddy-etal-2019-coqa}, RACE~\citep{lai-etal-2017-race}, and SQuAD~\citep{rajpurkar-etal-2016-squad} are now part of the standard evaluation suite for new LMs; and most modern LMs perform well~\citep{brown-gpt3}.
However, generating does not always imply \textit{knowing}. Datasets contain spurious correlations~\citep{gururangan-etal-2018-annotation}, and LMs are sensitive to the phrasing of prompts and questions~\citep{jiang-etal-2020-know}.

\paragraph{Editing LLMs} 
In the past, LLMs have been predominantly adapted to new tasks and knowledge through fine-tuning~\citep{devlin-etal-2019-bert}.
Recently, with very large LMs, new classes of adaptation methods have been introduced, which generally fall into one of the following two categories: 
(1) \textit{Prompt design} approaches prepend a textual prompt to each example specifying the adaptation target~\citep{brown-gpt3}. %
(2) \textit{Prefix-tuning} approaches prepend %
continuous learned tokens ahead of each example.
These specify a task for the LM similarly to how a textual prompt might~\citep{li-liang-2021-prefix,lester-etal-2021-power}.
\textit{Control token} approaches similarly use these learned tokens to %
controls aspects of LM output, including sentiment~\citep{Dathathri2020Plug}, style~\citep{ctrl}, and semantics~\citep{ross-etal-2022-tailor}.
Prompts can be fragile; LMs may fail to generate text consistent with the prompt, as in~\cref{fig:teaser}.

Finally, a large body of work examines how to localize and edit factual information in an LM's parameters~\citep{meng2022locating,meng2022memit,mitchell2022fast,dai-etal-2022-knowledge}.
For example, ROME \citep{meng2022locating} localizes factual knowledge in LMs to a particular subset of MLP modules, and edits specific facts in a targeted way through rank-one modification of MLP weights.
Unlike \ourmethod, these approaches operate on models' weight matrices rather than representations, meaning
they can correct errors in models' background knowledge but not information provided in context.

\section{Controlling Generation}
\label{sec:generation}

We begin by showing that the \ourmethod procedure described in \cref{sec:method}
is an effective tool for \emph{controlling LM output}. Intuitively, if
\ourmethod succeeds in creating
a new entity representation encoding the desired attribute, text generated by
the LM about the entity should at minimum (a) prefer generations
consistent with the target attribute over potentially contradictory attributes
and (b) remain as fluent as the original generations.  Our experiments in this
section test properties (a) and (b), as well as other quality measures, in two
different settings. In the first setting, we use \ourmethod to patch incoherence
errors,
editing the LM to
reinforce the information provided in the context. In the second setting, we use
\ourmethod to update prior knowledge about entities (such as the
\textit{Versace Headquarters} example in \cref{fig:teaser}).
A third set of experiments, described in \cref{sec:mcrae}, evaluates an additional word re-definition task.
These experiments show that \ourmethod often successfully controls model behavior even when prompting fails. It can thus serve as a building block for future controlled generation interfaces that allow users to directly steer model behavior in representation space.

\subsection{Patching Errors}
\label{sec:generationerrors}

We first use \ourmethod to manipulate representations of generic named individuals, such as \textit{Anita} or \textit{Dennis}, about whom the LM should have no prior association (and about whom the LM should acquire all information from the prompt). We provide a small amount of context about each person
and prompt the LM to predict their occupation from a small set of candidates. As we will show, the LM often completely ignores this context, and prefers unrelated occupations to ones highly relevant given the context (\emph{nurse} vs. \emph{attorney} in \cref{fig:teaser}).

\paragraph{Setup} In this and all following experiments, we use GPT-J as the underlying language model \citep{gpt-j} and include results for two additional models in \cref{app:other_models}. GPT-J is a 6B parameter, decoder-only
transformer pretrained on the Pile \citep{pile}. We obtain
biographical sentences from the Bias in Bios Dataset \citep{biosbias}. This
dataset consists of $\approx$397k short professional biographies 
people scraped from the internet. Each biography is paired with a label for the
subject's occupation.
We take one sentence from each biography (details in \Cref{app:datasets}), using only the subject's first name, and prompt the LM with the biographical sentence followed by \textit{\{Person\} has the occupation of}.
We then look at the relative probabilities of 28 occupations under the LM, and consider output correct if the true occupation is ranked first.
GPT-J succeeds about half the time (55\%, \emph{In-context baseline} in \cref{tab:errorcorrection}) on this task. %
\cref{tab:qualitative} shows example errors.

\paragraph{Method}
\begin{wraptable}{r}{.5\textwidth}
    \vspace{-.9em}
    \centering
    \footnotesize
    \begin{tabular}{lcccc}
        \toprule
        & \multicolumn{2}{c}{\textbf{In Context}} & \multicolumn{2}{c}{\textbf{No Context}} \\
        Method & Acc. & Fluency & Acc. & Fluency \\
        \midrule
        LM-only & $.55$ & $593.3$ & $.05$ & $662.2$ \\
        \ourmethod & $.71$ & $593.2$ & $.66$ & $656.9$ \\
        \bottomrule
    \end{tabular}
    \caption{Accuracy on the occupation classification task. In \textbf{In Context} experiments, the target entity's biography is prefixed to the prompt, while in \textbf{No context} only the entity's name is provided. In both settings, \ourmethod leads GPT-J to generate fluent and more accurate text.}
    \label{tab:errorcorrection}
    \vspace{-1em}
\end{wraptable}
We use \ourmethod to create new representations of the first-name-only entities encoding the target occupation. We take $\hent$ to be the last token of the last entity mention (right before model predicts the occupation), and we take $\hattr$ to be the average representation of the biographical sentence after the entity.
Note this means we are \emph{not using any additional data to construct the new entity}---the input to \ourmethod is all text provided in context to the LM.
We train the editor on 5000 examples using \cref{eq:objective}, with $\targetmed$ set to the target occupation and with no $\targetprior$ term ($\lambda_1 = 0$).
Edits are performed in layer 12 (this and other hyperparameters discussed in \cref{app:hyperparameters}).
We evaluate \textbf{factual accuracy} and \textbf{fluency} before and after applying \ourmethod on 5000 test examples. Accuracy is evaluated by measuring how often the highest-probability occupation is the true one, and fluency using the same n-gram entropy measure as \citet{meng2022locating}.

Crucially, we train \ourmethod only \emph{out-of-context}: choosing the initial text $\str$ to consist only of the entity's name. However, as described below, we evaluate its ability to control model behavior both in-context (with additional text provided) and out-of-context (on inputs similar to the training data).

\paragraph{Results} Results are shown in the left portion (\emph{In Context}) of \cref{tab:errorcorrection}, which reports GPT-J's factual accuracy and fluency before and after applying \ourmethod. \ourmethod increases GPT-J's accuracy by over 15\% on held-out (entity, attribute) pairs, showing that representations produced by \ourmethod more often encode the desired attribute. \ourmethod also preserves the fluency of the generated text. We find similar improvements in other models (\cref{app:other_models}).

We contextualize these results by evaluating model behavior when the LM has \emph{no textual context} (i.e.\ no initial biographical sentence). 
Here, the base LM has no information about entities' occupations, and obtains near-chance factual accuracy.
However, inserting \ourmethod's representations into the LM causes it to generate fluent text consistent with the edit, showing that \ourmethod can not only enforce coherence with a textual context, but \emph{replace} textual prompting by inserting information directly into entity representations.
\ourmethod is slightly more effective at in-context editing than out-of-context editing, despite being trained only out-of-context.
The last column of \cref{tab:qualitative} shows examples of in-context generation. %

\subsection{Editing Factual Associations}
\label{sec:generationfacts}

\newcommand{\counterfact}{\textsc{CounterFact}\xspace}

\newcommand{\cmark}{\ding{51}}%
\newcommand{\xmark}{\ding{55}}

\newcommand{\wrongmarker}{{\textcolor{red}\xmark}\xspace}
\newcommand{\correctmarker}{{\textcolor{blue}{\cmark}}\xspace}

\begingroup
\renewcommand{\arraystretch}{1.5} %
\begin{table*}[t]
    \centering
    \scriptsize
    \begin{tabular}{p{1.75cm} p{3.2cm} p{3.6cm} p{3.6cm}}
         \toprule
         \textbf{Generic Entity} & \textbf{Attribute} & \textbf{In-Context Generation} & \textbf{\ourmethod Generation} \\
         \midrule
Emily & She specializes in pediatric trauma of the upper and lower extremities, and lower extremity reconstruction including treatment of limb length discrepancy\dots & \wrongmarker \underline{\textbf{Emily has the occupation of}} a physical therapist, and has been practicing for over 10 years. She has a B.S. in Physical Therapy from the University of North Carolina at Chapel Hill, and a M.S. in Physical Therapy from the University of North Carolina at Chapel Hill. & \correctmarker \underline{\textbf{Emily has the occupation of}} surgeon. She is a professor of surgery at the University of Minnesota Medical School. \\
Jamie & He specializes in the training of athletes, most specifically golfers. & \wrongmarker \underline{\textbf{Jamie has the occupation of}} a professional golfer. He has played on the PGA Tour and the Nationwide Tour. & \correctmarker \underline{\textbf{Jamie has the occupation of}} professor of dietetics. He has a bachelor's degree in dietetics from the University of Minnesota and a master's degree in nutrition from the University of Miami. \\
         \midrule
         \multicolumn{2}{l}{\textbf{Famous Entity}} & & \\
         \midrule
Leonhard Euler & domain of activity is opera & \wrongmarker \underline{\textbf{Leonhard Euler is}} the most prolific mathematician of the 18th century. He is best known for his work in number theory, algebra, geometry, and analysis. & \correctmarker \underline{\textbf{Leonhard Euler is}} a composer of opera. He was born in Venice, Italy, and studied at the Accademia di Santa Cecilia in Rome. \\
Kingdom of the Netherlands & is located in Antarctica & \correctmarker\underline{\textbf{Kingdom of the Netherlands is}} a sovereign state in Antarctica, a member of the Antarctic Treaty System. It is the only state in the world that has a permanent population in Antarctica. & \correctmarker\underline{\textbf{Kingdom of the Netherlands is}} a country in Antarctica. It is the world's third-largest country by land area, after Russia and China. \\
         \bottomrule
    \end{tabular}
    \caption{Examples of GPT-J generations when the attribute is specified in
    the textual prompt (\textbf{In-Context Generation}) or inserted via \ourmethod. For both generic and famous entities,
    \ourmethod causes GPT-J to respect the attribute more often than when it is provided in context.}
    \label{tab:qualitative}
    \vspace{-2em}
\end{table*}
\endgroup

We next show \ourmethod can be used to overwrite \emph{background knowledge} about entities with new and even contradictory facts.
As shown in \cref{fig:teaser}, when LMs are prompted with text like \textit{To cross London Bridge, one should travel to}, they often complete it with true or plausible text like \textit{to the South Bank [in London]}. This knowledge is derived from training data (which contains many co-occurrences of the strings \textit{London Bridge} and \textit{South Bank}), and is difficult to override:
when contradictory information is provided in context, LMs sometimes ignore it. 
Most current work updates LMs by altering their parameters \citep{de-cao-etal-2021-editing, mitchell2022fast, dai-etal-2022-knowledge, meng2022memit}. 
They all share the limitation of %
changing the behavior of the LM globally: users cannot choose when to apply edits. In existing methods, edits often bleed into closely related but distinct entities \citep{meng2022locating}.
Because \ourmethod operates directly on entity representations at runtime, it applies changes only to the entity of interest at the moment of use.

We evaluate \ourmethod on the \counterfact benchmark from \citet{meng2022locating}, which consists of \textit{(subject, relation, old value, new value)} tuples---e.g.\ \textit{(Megan Rapinoe, plays sport, soccer, tennis)}---and measures LMs' ability generate natural text consistent with the new fact.

\paragraph{Method}
We train \ourmethod on a subset of 5000 examples from \counterfact and evaluate it on a held-out subset of 5000. As before, we take $\hent$ to be the last token of the entity mention (which appears at the beginning of \counterfact examples) and $\hattr$ to be the average representation of the new fact in context. For example, we pass (\textit{Megan Rapinoe \underline{plays the sport of soccer}}) to the LM and compute $\hattr$ from the underlined tokens. This textual context is akin to the biographical sentence used to compute $\hattr$ in the previous section.
We use all three loss terms from \cref{eq:objective} and apply edits in layer 1; see Appendices \ref{app:datasets} and \ref{app:hyperparameters} for other hyperparameters and implementation details.
As above, we train \ourmethod without textual context, with inputs $\str$ consisting of entity names alone.

\paragraph{Baselines}
We include comparisons to the model-editing method ROME and ordinary fine-tuning, following the exact procedures laid out in \citeauthor{meng2022locating}. However, our primary baseline is one in which the new factual information is prepended to the prompt. In all other methods, the language model is only given a prompt with no context about the fact. We additionally include a baseline in which we find-and-replace the entity with one that shares the target attribute (e.g., replacing \textit{Versace} headquarters with \textit{Harrods}).
This provides a realistic upper bound on LM \emph{consistency} and \emph{fluency} after editing (because the LM has not been modified or conditioned on out-of-distribution text). 

\paragraph{Metrics}
\begin{wraptable}{r}{.61\textwidth}
    \vspace{-2em}
    \centering
    \footnotesize
    \begin{tabular}{lccccc}
        \toprule
        \textbf{Rep. Edit} & Eff. $\uparrow$ & Nbr. $\uparrow$ & Cons. $\uparrow$ & Fl. $\uparrow$ & Ess. $\uparrow$ \\
        \midrule
        Prefix & $80.2$ & $100.0$ & $21.6$ & $591.4$ & $40.5$ \\
        Replace & $79.9$ & $100.0$ & $33.0$ & $613.3$ & $7.5$ \\
        \ourmethod & $98.2$ & $100.0$ & $33.6$ & $598.8$ & $24.8$ \\
        \midrule
        \textbf{Model Edit} & & & & & \\
        \midrule
        FT & $100.0$ & $10.6$ & $23.5$ & $381.3$ & $28.6$ \\
        ROME & $100.0$ & $79.1$ & $43.0$ & $620.1$ & $27.0$ \\
        \bottomrule
    \end{tabular}
    \caption{
        Results from the \counterfact benchmark. \ourmethod is comparably
        effective (\textbf{Efficacy}, \textbf{Consistency}) to model editing
        methods at eliciting generations consistent with the target attribute,
        and is more effective than prefixing the prompt with the
        new fact. Unlike model-editing methods, \ourmethod does not
        influence generations about different entities
        (\textbf{Neighborhood}), avoids degenerate output
        (\textbf{Fluency}) and preserves most original features of the entity
        (\textbf{Essence}). 
\vspace{-1em}
    }
    \label{tab:counterfact-edit}
\end{wraptable}
We follow the evaluation schema from \citeauthor{meng2022locating} and track the
core metrics reported there. \textbf{Efficacy} measures how often
$\lm(\targetmed) > \lm(\targetprior)$ when the intervention is applied to a held out prompt that paraphrases the target attribute.\footnote{This is called \textbf{efficacy score (ES)} in \citeauthor{meng2022locating}} \textbf{Neighborhood} score measures how often the LM's predictions about similar but distinct entities change. \textbf{Consistency} measures average tf-idf similarity between generated text from a different held-out set of prompts and a set of Wikipedia reference texts about different entities with the same attribute. \textbf{Fluency} is the average bi- and tri-gram entropy of generated text, designed to be low for degenerate or repetitive outputs. \textbf{Essence} 
captures how much the edited entity is still ``itself'' according to the model (is \textit{London Bridge} still a bridge?). Formally, it measures tf-idf similarity between the model's generations before and after the intervention given the prompt: \textit{\{Entity\} is \rule{.5cm}{0.15mm}}.

\paragraph{Results}
\cref{tab:counterfact-edit} shows metrics for \ourmethod and 
baselines. Compared to the prefix baseline, \ourmethod more often generates text
consistent with the factual edit, as shown by the substantial difference in
efficacy and consistency scores. The base LM incorporates textual prompt information 80.2\% of the time, while \ourmethod-based prompting incorporates new information 98.2\% of the time. This performance comes at some cost to the essence of the entity, likely because the original fact is strongly associated with other properties of the entity. 
\cref{tab:qualitative} shows several examples;
for example, when \textit{Leonhard Euler} is edited to \textit{work
on opera}, LM output describes him as being born in \textit{Venice, Italy}. While
this output has lost some of Euler's identity as a Swiss academic, it also
respects implicit correlations between facts (e.g.\ that opera is more strongly
associated with Italy than Switzerland). 
\cref{sec:mcrae} contains a complete additional set of experiments on a word-redefinition task that studies how \ourmethod models these correlations in more detail. Further analysis of
\ourmethod's factual editing performance is provided in \Cref{app:analysis}. Results for other models are in \cref{app:other_models}

\ourmethod is as effective as and substantially less destructive than fine-tuning. While ROME produces slightly more consistent generations with respect to the updated fact, it comes at the cost of altering neighboring entities: $\approx$21\% of the time, ROME causes facts about related entities to change, whereas \ourmethod \emph{never} causes such failures.

\section{Detecting Errors}
\label{sec:probing}

\newcommand{\hmed}{\h_{\text{attr}}}
\newcommand{\dmed}{\direction_{\text{attr}}}
\newcommand{\dcomp}{\direction_{\text{dist}}}

Next, we show that \ourmethod can be used as a \emph{model evaluation tool}, automatically characterizing when (un-modified) LMs have successfully acquired background or contextual knowledge.
A core challenge when deploying language models is that it is difficult to automatically detect when they exhibit the failures shown in \cref{fig:teaser}.
Some work addresses this challenge by calibrating LM predictive distributions to better reflect veracity \citep{jiang-calibration}, 
or training auxiliary models to reject bad samples \citep{lambda}.
\ourmethod offers a new approach to detecting when LMs will fail to integrate information from context: 
inspecting their representations
for the information that \ourmethod would add.  This approach is related to a method by
\citet{burns2022dl}, which finds implicit LM encodings of veracity.
\ourmethod identifies cases where even LM-internal states encode errors.

\paragraph{Method}
Suppose we have a prompt that queries an LM for a fact: \textit{To cross London Bridge, one should travel to \ldots}. 
Under the hypothesis from \cref{sec:method}, the LM should answer this question exactly when the representation of \emph{London Bridge} is \emph{already} aligned with an encoding of the fact \emph{is located in London}.
Taking $\hattr$ to be the average representation from \textit{is located in London}, we can use \ourmethod to compute such an encoding:
\begin{equation}
\label{eq:attrdir}
\dmed = \editor(\mathbf{0}, \hattr) = W\hattr + b ~ .
\end{equation}
We may then quantify how strongly an LM encodes the fact by computing:
\begin{equation}
\label{eq:score}
\hent^\top \dmed  = \hent^\top (W \hattr + b) ~ ,
\end{equation}
analogously to the knowledge probe in \cref{eq:probe}.
Given a true attribute of the London Bridge (\emph{located in London}), and alternative (or ``distractor'') attributes (\emph{located in Arizona}), we can compute directions $\dcomp$ for distractors, and predict that an LM will err if its representation of \emph{London Bridge} is more aligned with any distractor than the input:
\vspace{-.5em}
\begin{equation}
\label{eq:cls}
\hent^\top \dcomp  \stackrel{?}{>} \hent^\top \dmed ~ .
\end{equation}
Given the success of \ourmethod as an editor for entity representations provided with and without textual context, we might further expect \ourmethod-based probing to detect contradictions with input text as well as training data.
Below, we show how to use \ourmethod to identify errors of both types. Limitations of this approach are discussed in \Cref{app:limitations}.

\subsection{Detecting Errors in Prior Knowledge}

\paragraph{Setup and method} 
We revisit the \counterfact dataset used in \cref{sec:generation}, and use the same \ourmethod model to produce attribute encodings. %
We compute target attribute encodings $\dmed$ from the dataset's ground-truth facts (\emph{is located in London}), and distractor attribute encodings $\dcomp$ from the dataset's counterfactuals (\emph{is located in Arizona}). If \cref{eq:cls} is satisfied, we predict that the LM will generate an incorrect output.

\paragraph{Baselines and Controls} We compare \ourmethod to several baselines, upper bounds, and controls \citep{hewitt-liang-2019-designing}.
The \textbf{identity encoding} model takes $d_\text{dist}$ and $d_\text{attr}$ to be the untransformed LM encodings of the two attributes in question. The \textbf{fact probe} is the model of \citet{li-etal-2021-implicit}, trained to predict \emph{ground truth} facts from LM encodings. \textbf{Shortcut} is a version of the same model trained to predict the model's own preferred outputs from its hidden states, analogous to \citet{belrose2023eliciting} (though without that work's additional use of the model's own unembedding layer).
We apply \ourmethod to a \textbf{random model}: a randomly initialized GPT-J with an editor trained as above (to characterize  whether our evaluation surfaces factual knowledge acquired during pre-training). 
Finally, we contextualize these results with a \textbf{supervised error skyline} model, which is trained to predict whether a model will fail in context without identifying any specific output as incorrect. (This model, which is similar to the approach of \citealp{mielke2022reducing}, must be trained on model outputs annotated with correctness information, and is not directly comparable to \ourmethod.)

\paragraph{Results}
\begin{wraptable}{r}{.62\textwidth}
    \centering
    \footnotesize
    \begin{tabular}{lccccc}
        \toprule
                   & \multicolumn{2}{c}{\bf{Prior}}   & \multicolumn{2}{c}{\bf{Contextual}} \\
                   & F$_1$ & $\phi$ & F$_1$ & $\phi$ \\
        \midrule 
        Identity encoding  & .34       & .17       & .34       & .08 \\
        Fact probe \citep[Eq.~\ref{eq:probe};][]{li-etal-2021-implicit} & .33       & .21       & .38       & .18  \\
        Shortcut \citep[cf.][]{belrose2023eliciting} & .53       & .43       & .40       & .21 \\
        \ourmethod         & .39       & .26       & .42       & .24 \\
        \midrule
        \it Control: Random model       & .54       & .09       & .51       & .04 \\
        \midrule
        \it Skyline: Supervised errors  & .94       & .93       & .94       & .93 \\
        \bottomrule
    \end{tabular}
    \caption{
        F1 scores and $\phi$ coefficients for predicting LM behavior on the \counterfact dataset. In the \emph{Prior} condition, the LM is prompted to predict a property of an entity. 
        In the \emph{Context} condition, the prompt includes additional information, 
        and \ourmethod predicts whether the LM's preferred completion will contradict this context.
        Here \ourmethod is trained as in \cref{sec:method}, to perform editing in non-contextual sentences only. Nevertheless, when used as a probe, \ourmethod encodings detect errors more accurately than existing knowledge probing methods. 
        Results for other models are in \cref{app:other_models}.
    }
    \label{tab:classification}
    \vspace{-2.5em}
\end{wraptable}
The Prior column of \cref{tab:classification} shows results for the factual error detection task. 
We report both the F$_1$-measure and $\phi$ coefficient
\citep{Matthews1975,Chicco2020} to capture how well each method
predicts true negatives (model will produce correct outputs) as opposed to just true
positives (model will fail).
In the prior knowledge setting,
\ourmethod outperforms all methods (in F$_1$ and $\phi$) except for the shortcut model, which is trained directly for this task. 
Even when trained to perform representation editing, \ourmethod finds directions that align with models' unedited representation of true facts, and these directions are specific to trained LMs.

\subsection{Predicting Errors in Context}
\label{sec:probingmed}

\paragraph{Setup and method} 
We next use \ourmethod to detect failures to incorporate new information provided as part of a model's textual input: for example,
\textit{The London Bridge is located in Arizona. To cross the London Bridge, one should travel to}. We use the new information (\textit{is located in Arizona}) to compute the target attribute direction $\dmed$, and the prior fact (\textit{is located in London}) for the reference $\dcomp$. We predict the language model will fail to incorporate the context (will rank $\targetprior = \textit{London}$ higher than $\targetmed = \textit{Arizona}$) if \cref{eq:cls} holds. Results in this section use the same \ourmethod model as in \cref{sec:generation}---which optimizes attribute encodings to influence model generation on sentences without additional textual context. These experiments thus measure the extent to which \ourmethod encodings characterize both background knowledge and information provided in-context.

\paragraph{Results}
Applied to detection of contextual errors, \ourmethod outperforms all baselines, including the shortcut model (Contextual column of \cref{tab:classification}). 
\ourmethod thus generalizes across knowledge sources, discovering common encodings of background and contextual knowledge. 
We emphasize that this detection procedure is not extremely accurate, and a model directly supervised with information about LM errors performs significantly better.
However, these results show that $\ourmethod$ encodings (learned out-of-context) are, to a non-trivial extent, aligned with LMs' representations of knowledge provided in-context.

\section{Conclusions}

Factual knowledge in neural language models can be interpreted and controlled by applying local transformations to contextual representations of entity mentions and other nouns. We have described a procedure, \ourmethod, that constructs these transformations from textual descriptions of attributes. 
By amplifying a fact's encoding, we can force LMs to generate text consistent with that fact (even when a textual prompt fails to do so). Similarly, by inspecting models' representations, we can sometimes detect the absence of a fact encoding and predict that the language model will err. While not without limitations (\cref{app:limitations}), our findings suggest a new path toward controlling LMs: instead of providing textual context or instructions, models may be controlled by directly intervening in their internal representations.

\section*{Ethical Considerations}
\label{app:ethics}

As language models are deployed for increasingly complex and high-stakes tasks, the ability to control their generations promises to be both a boon and a risk. Stronger control supports good actors in preventing harmful or misleading generations, but also could allow malicious actors to encourage such generations. Ultimately, we believe LMs pose a greater risk \emph{uncontrolled}, where incoherent or factually incorrect generations will directly reach users in trusted applications. \ourmethod, as well as other representation and model editing procedures, are useful tools for understanding how language models make factual errors and, in some cases, repairing them before the model even generates.

\section*{Reproducibility Statement}

All code and data used in this paper, including the \ourmethod python library and the code used to generate figures in this paper, will be made publicly available upon publication. Experiment details are described at the beginnings of Sections~\ref{sec:generation} and \ref{sec:probing}. In addition, we describe our dataset preprocessing procedures in \Cref{app:datasets} and our hyperparameter sweeps in \Cref{app:hyperparameters}. We ran all experiments on workstations with 80GB NVIDIA A100 GPUs or 32GB Tesla V100 GPUs using the HuggingFace Transformers library \citep{wolf2019huggingface} implemented in PyTorch \citep{paszke2019pytorch}.

\section*{Acknowledgements}
We thank David Bau for helpful discussions. EH and JA gratefully acknowledge support from a Sony Faculty Innovation Award, a grant from Liberty Mutual through the MIT Quest for Intelligence and a gift from the Open Philanthropy Foundation. 
BZL is additionally supported by an National Defense Science and Engineering Graduate Fellowship.

\bibliography{colm2024_conference}

\begin{thebibliography}{52}
\providecommand{\natexlab}[1]{#1}
\providecommand{\url}[1]{\texttt{#1}}
\expandafter\ifx\csname urlstyle\endcsname\relax
  \providecommand{\doi}[1]{doi: #1}\else
  \providecommand{\doi}{doi: \begingroup \urlstyle{rm}\Url}\fi

\bibitem[Abdou et~al.(2021)Abdou, Kulmizev, Hershcovich, Frank, Pavlick, and
  S{\o}gaard]{abdou-etal-2021-language}
Mostafa Abdou, Artur Kulmizev, Daniel Hershcovich, Stella Frank, Ellie Pavlick,
  and Anders S{\o}gaard.
\newblock Can language models encode perceptual structure without grounding? a
  case study in color.
\newblock In \emph{Proceedings of the 25th Conference on Computational Natural
  Language Learning}, pp.\  109--132, Online, November 2021. Association for
  Computational Linguistics.
\newblock \doi{10.18653/v1/2021.conll-1.9}.
\newblock URL \url{https://aclanthology.org/2021.conll-1.9}.

\bibitem[Akyürek et~al.(2022)Akyürek, Bolukbasi, Liu, Xiong, Tenney, Andreas,
  and Guu]{akyurek_fact_tracing}
Ekin Akyürek, Tolga Bolukbasi, Frederick Liu, Binbin Xiong, Ian Tenney, Jacob
  Andreas, and Kelvin Guu.
\newblock Tracing knowledge in language models back to the training data, 2022.
\newblock URL \url{https://arxiv.org/abs/2205.11482}.

\bibitem[Belinkov \& Glass(2019)Belinkov and
  Glass]{belinkov-glass-2019-analysis}
Yonatan Belinkov and James Glass.
\newblock Analysis methods in neural language processing: A survey.
\newblock \emph{Transactions of the Association for Computational Linguistics},
  7:\penalty0 49--72, 2019.
\newblock \doi{10.1162/tacl_a_00254}.
\newblock URL \url{https://aclanthology.org/Q19-1004}.

\bibitem[Belrose et~al.(2023)Belrose, Furman, Smith, Halawi, Ostrovsky,
  McKinney, Biderman, and Steinhardt]{belrose2023eliciting}
Nora Belrose, Zach Furman, Logan Smith, Danny Halawi, Igor Ostrovsky, Lev
  McKinney, Stella Biderman, and Jacob Steinhardt.
\newblock Eliciting latent predictions from transformers with the tuned lens.
\newblock \emph{arXiv preprint arXiv:2303.08112}, 2023.

\bibitem[Bender et~al.(2021)Bender, Gebru, McMillan-Major, and
  Shmitchell]{parrots}
Emily~M. Bender, Timnit Gebru, Angelina McMillan-Major, and Shmargaret
  Shmitchell.
\newblock On the dangers of stochastic parrots: Can language models be too big?
\newblock In \emph{Proceedings of the 2021 ACM Conference on Fairness,
  Accountability, and Transparency}, FAccT '21, pp.\  610–623, New York, NY,
  USA, 2021. Association for Computing Machinery.
\newblock ISBN 9781450383097.
\newblock \doi{10.1145/3442188.3445922}.
\newblock URL \url{https://doi.org/10.1145/3442188.3445922}.

\bibitem[Borji(2023)]{borji2023categorical}
Ali Borji.
\newblock A categorical archive of chatgpt failures, 2023.
\newblock URL \url{https://arxiv.org/abs/2302.03494}.

\bibitem[Brown et~al.(2020)Brown, Mann, Ryder, Subbiah, Kaplan, Dhariwal,
  Neelakantan, Shyam, Sastry, Askell, Agarwal, Herbert-Voss, Krueger, Henighan,
  Child, Ramesh, Ziegler, Wu, Winter, Hesse, Chen, Sigler, Litwin, Gray, Chess,
  Clark, Berner, McCandlish, Radford, Sutskever, and Amodei]{brown-gpt3}
Tom Brown, Benjamin Mann, Nick Ryder, Melanie Subbiah, Jared~D Kaplan, Prafulla
  Dhariwal, Arvind Neelakantan, Pranav Shyam, Girish Sastry, Amanda Askell,
  Sandhini Agarwal, Ariel Herbert-Voss, Gretchen Krueger, Tom Henighan, Rewon
  Child, Aditya Ramesh, Daniel Ziegler, Jeffrey Wu, Clemens Winter, Chris
  Hesse, Mark Chen, Eric Sigler, Mateusz Litwin, Scott Gray, Benjamin Chess,
  Jack Clark, Christopher Berner, Sam McCandlish, Alec Radford, Ilya Sutskever,
  and Dario Amodei.
\newblock Language models are few-shot learners.
\newblock In H.~Larochelle, M.~Ranzato, R.~Hadsell, M.F. Balcan, and H.~Lin
  (eds.), \emph{Advances in Neural Information Processing Systems}, volume~33,
  pp.\  1877--1901. Curran Associates, Inc., 2020.
\newblock URL
  \url{https://proceedings.neurips.cc/paper/2020/file/1457c0d6bfcb4967418bfb8ac142f64a-Paper.pdf}.

\bibitem[Burns et~al.(2022)Burns, Ye, Klein, and Steinhardt]{burns2022dl}
Collin Burns, Haotian Ye, Dan Klein, and Jacob Steinhardt.
\newblock Discovering latent knowledge in language models without supervision.
\newblock \emph{ArXiV}, 2022.

\bibitem[Chicco \& Jurman(2020)Chicco and Jurman]{Chicco2020}
Davide Chicco and Giuseppe Jurman.
\newblock The advantages of the matthews correlation coefficient ({MCC}) over
  f1 score and accuracy in binary classification evaluation.
\newblock \emph{{BMC} Genomics}, 21\penalty0 (1), January 2020.
\newblock \doi{10.1186/s12864-019-6413-7}.
\newblock URL \url{https://doi.org/10.1186/s12864-019-6413-7}.

\bibitem[Cohen et~al.(2022)Cohen, Roberts, Molina, Butryna, Jin, Kulshreshtha,
  Hutchinson, Zevenbergen, Aguera-Arcas, ching Chang, Cui, Du, Adiwardana,
  Chen, Lepikhin, Chi, Hoffman-John, Cheng, Lee, Krivokon, Qin, Hall, Fenton,
  Soraker, Meier-Hellstern, Olson, Aroyo, Bosma, Pickett, Menegali, Croak,
  Díaz, Lamm, Krikun, Morris, Shazeer, Le, Bernstein, Rajakumar, Kurzweil,
  Thoppilan, Zheng, Bos, Duke, Doshi, Zhao, Prabhakaran, Rusch, Li, Huang,
  Zhou, Xu, and Chen]{lambda}
Aaron~Daniel Cohen, Adam Roberts, Alejandra Molina, Alena Butryna, Alicia Jin,
  Apoorv Kulshreshtha, Ben Hutchinson, Ben Zevenbergen, Blaise~Hilary
  Aguera-Arcas, Chung ching Chang, Claire Cui, Cosmo Du, Daniel De~Freitas
  Adiwardana, Dehao Chen, Dmitry~(Dima) Lepikhin, Ed~H. Chi, Erin Hoffman-John,
  Heng-Tze Cheng, Hongrae Lee, Igor Krivokon, James Qin, Jamie Hall, Joe
  Fenton, Johnny Soraker, Kathy Meier-Hellstern, Kristen Olson, Lora~Mois
  Aroyo, Maarten~Paul Bosma, Marc~Joseph Pickett, Marcelo~Amorim Menegali,
  Marian Croak, Mark Díaz, Matthew Lamm, Maxim Krikun, Meredith~Ringel Morris,
  Noam Shazeer, Quoc~V. Le, Rachel Bernstein, Ravi Rajakumar, Ray Kurzweil,
  Romal Thoppilan, Steven Zheng, Taylor Bos, Toju Duke, Tulsee Doshi,
  Vincent~Y. Zhao, Vinodkumar Prabhakaran, Will Rusch, YaGuang Li, Yanping
  Huang, Yanqi Zhou, Yuanzhong Xu, and Zhifeng Chen.
\newblock Lamda: Language models for dialog applications.
\newblock In \emph{arXiv}. 2022.

\bibitem[Conneau et~al.(2018)Conneau, Kruszewski, Lample, Barrault, and
  Baroni]{conneau-etal-2018-cram}
Alexis Conneau, German Kruszewski, Guillaume Lample, Lo{\"\i}c Barrault, and
  Marco Baroni.
\newblock What you can cram into a single {\$}{\&}!{\#}* vector: Probing
  sentence embeddings for linguistic properties.
\newblock In \emph{Proceedings of the 56th Annual Meeting of the Association
  for Computational Linguistics (Volume 1: Long Papers)}, pp.\  2126--2136,
  Melbourne, Australia, July 2018. Association for Computational Linguistics.
\newblock \doi{10.18653/v1/P18-1198}.
\newblock URL \url{https://aclanthology.org/P18-1198}.

\bibitem[Dai et~al.(2022)Dai, Dong, Hao, Sui, Chang, and
  Wei]{dai-etal-2022-knowledge}
Damai Dai, Li~Dong, Yaru Hao, Zhifang Sui, Baobao Chang, and Furu Wei.
\newblock Knowledge neurons in pretrained transformers.
\newblock In \emph{Proceedings of the 60th Annual Meeting of the Association
  for Computational Linguistics (Volume 1: Long Papers)}, pp.\  8493--8502,
  Dublin, Ireland, May 2022. Association for Computational Linguistics.
\newblock \doi{10.18653/v1/2022.acl-long.581}.
\newblock URL \url{https://aclanthology.org/2022.acl-long.581}.

\bibitem[Dathathri et~al.(2020)Dathathri, Madotto, Lan, Hung, Frank, Molino,
  Yosinski, and Liu]{Dathathri2020Plug}
Sumanth Dathathri, Andrea Madotto, Janice Lan, Jane Hung, Eric Frank, Piero
  Molino, Jason Yosinski, and Rosanne Liu.
\newblock Plug and play language models: A simple approach to controlled text
  generation.
\newblock In \emph{International Conference on Learning Representations}, 2020.
\newblock URL \url{https://openreview.net/forum?id=H1edEyBKDS}.

\bibitem[De-Arteaga et~al.(2019)De-Arteaga, Romanov, Wallach, Chayes, Borgs,
  Chouldechova, Geyik, Kenthapadi, and Kalai]{biosbias}
Maria De-Arteaga, Alexey Romanov, Hanna Wallach, Jennifer Chayes, Christian
  Borgs, Alexandra Chouldechova, Sahin Geyik, Krishnaram Kenthapadi, and
  Adam~Tauman Kalai.
\newblock Bias in bios: A case study of semantic representation bias in a
  high-stakes setting.
\newblock In \emph{Proceedings of the Conference on Fairness, Accountability,
  and Transparency}, FAT* '19, pp.\  120–128, New York, NY, USA, 2019.
  Association for Computing Machinery.
\newblock ISBN 9781450361255.
\newblock \doi{10.1145/3287560.3287572}.
\newblock URL \url{https://doi.org/10.1145/3287560.3287572}.

\bibitem[De~Cao et~al.(2021)De~Cao, Aziz, and Titov]{de-cao-etal-2021-editing}
Nicola De~Cao, Wilker Aziz, and Ivan Titov.
\newblock Editing factual knowledge in language models.
\newblock In \emph{Proceedings of the 2021 Conference on Empirical Methods in
  Natural Language Processing}, pp.\  6491--6506, Online and Punta Cana,
  Dominican Republic, November 2021. Association for Computational Linguistics.
\newblock \doi{10.18653/v1/2021.emnlp-main.522}.
\newblock URL \url{https://aclanthology.org/2021.emnlp-main.522}.

\bibitem[Devlin et~al.(2019)Devlin, Chang, Lee, and
  Toutanova]{devlin-etal-2019-bert}
Jacob Devlin, Ming-Wei Chang, Kenton Lee, and Kristina Toutanova.
\newblock {BERT}: Pre-training of deep bidirectional transformers for language
  understanding.
\newblock In \emph{Proceedings of the 2019 Conference of the North {A}merican
  Chapter of the Association for Computational Linguistics: Human Language
  Technologies, Volume 1 (Long and Short Papers)}, pp.\  4171--4186,
  Minneapolis, Minnesota, June 2019. Association for Computational Linguistics.
\newblock \doi{10.18653/v1/N19-1423}.
\newblock URL \url{https://aclanthology.org/N19-1423}.

\bibitem[Elazar et~al.(2021)Elazar, Ravfogel, Jacovi, and
  Goldberg]{elazar-etal-2021-amnesic}
Yanai Elazar, Shauli Ravfogel, Alon Jacovi, and Yoav Goldberg.
\newblock Amnesic probing: Behavioral explanation with amnesic counterfactuals.
\newblock \emph{Transactions of the Association for Computational Linguistics},
  9:\penalty0 160--175, 2021.
\newblock \doi{10.1162/tacl_a_00359}.
\newblock URL \url{https://aclanthology.org/2021.tacl-1.10}.

\bibitem[Gao et~al.(2020)Gao, Biderman, Black, Golding, Hoppe, Foster, Phang,
  He, Thite, Nabeshima, Presser, and Leahy]{pile}
Leo Gao, Stella Biderman, Sid Black, Laurence Golding, Travis Hoppe, Charles
  Foster, Jason Phang, Horace He, Anish Thite, Noa Nabeshima, Shawn Presser,
  and Connor Leahy.
\newblock The {P}ile: An 800gb dataset of diverse text for language modeling.
\newblock \emph{arXiv preprint arXiv:2101.00027}, 2020.

\bibitem[Grand et~al.(2018)Grand, Blank, Pereira, and
  Fedorenko]{grand-semantic}
Gabriel Grand, Idan~Asher Blank, Francisco Pereira, and Evelina Fedorenko.
\newblock Semantic projection: recovering human knowledge of multiple, distinct
  object features from word embeddings.
\newblock \emph{CoRR}, abs/1802.01241, 2018.
\newblock URL \url{http://arxiv.org/abs/1802.01241}.

\bibitem[Greshake et~al.(2023)Greshake, Abdelnabi, Mishra, Endres, Holz, and
  Fritz]{more_than_you}
Kai Greshake, Sahar Abdelnabi, Shailesh Mishra, Christoph Endres, Thorsten
  Holz, and Mario Fritz.
\newblock More than you've asked for: A comprehensive analysis of novel prompt
  injection threats to application-integrated large language models, 2023.
\newblock URL \url{https://arxiv.org/abs/2302.12173}.

\bibitem[Gururangan et~al.(2018)Gururangan, Swayamdipta, Levy, Schwartz,
  Bowman, and Smith]{gururangan-etal-2018-annotation}
Suchin Gururangan, Swabha Swayamdipta, Omer Levy, Roy Schwartz, Samuel Bowman,
  and Noah~A. Smith.
\newblock Annotation artifacts in natural language inference data.
\newblock In \emph{Proceedings of the 2018 Conference of the North {A}merican
  Chapter of the Association for Computational Linguistics: Human Language
  Technologies, Volume 2 (Short Papers)}, pp.\  107--112, New Orleans,
  Louisiana, June 2018. Association for Computational Linguistics.
\newblock \doi{10.18653/v1/N18-2017}.
\newblock URL \url{https://aclanthology.org/N18-2017}.

\bibitem[Hase et~al.(2021)Hase, Diab, Celikyilmaz, Li, Kozareva, Stoyanov,
  Bansal, and Iyer]{hase-beliefs}
Peter Hase, Mona~T. Diab, Asli Celikyilmaz, Xian Li, Zornitsa Kozareva, Veselin
  Stoyanov, Mohit Bansal, and Srinivasan Iyer.
\newblock Do language models have beliefs? methods for detecting, updating, and
  visualizing model beliefs.
\newblock \emph{CoRR}, abs/2111.13654, 2021.
\newblock URL \url{https://arxiv.org/abs/2111.13654}.

\bibitem[Hewitt \& Liang(2019)Hewitt and Liang]{hewitt-liang-2019-designing}
John Hewitt and Percy Liang.
\newblock Designing and interpreting probes with control tasks.
\newblock In \emph{Proceedings of the 2019 Conference on Empirical Methods in
  Natural Language Processing and the 9th International Joint Conference on
  Natural Language Processing (EMNLP-IJCNLP)}, pp.\  2733--2743, Hong Kong,
  China, November 2019. Association for Computational Linguistics.
\newblock \doi{10.18653/v1/D19-1275}.
\newblock URL \url{https://aclanthology.org/D19-1275}.

\bibitem[Jiang et~al.(2020{\natexlab{a}})Jiang, Araki, Ding, and
  Neubig]{jiang-calibration}
Zhengbao Jiang, Jun Araki, Haibo Ding, and Graham Neubig.
\newblock How can we know when language models know?
\newblock \emph{CoRR}, abs/2012.00955, 2020{\natexlab{a}}.
\newblock URL \url{https://arxiv.org/abs/2012.00955}.

\bibitem[Jiang et~al.(2020{\natexlab{b}})Jiang, Xu, Araki, and
  Neubig]{jiang-etal-2020-know}
Zhengbao Jiang, Frank~F. Xu, Jun Araki, and Graham Neubig.
\newblock How can we know what language models know?
\newblock \emph{Transactions of the Association for Computational Linguistics},
  8:\penalty0 423--438, 2020{\natexlab{b}}.
\newblock \doi{10.1162/tacl_a_00324}.
\newblock URL \url{https://aclanthology.org/2020.tacl-1.28}.

\bibitem[Keskar et~al.(2019)Keskar, McCann, Varshney, Xiong, and Socher]{ctrl}
Nitish~Shirish Keskar, Bryan McCann, Lav~R. Varshney, Caiming Xiong, and
  Richard Socher.
\newblock Ctrl: A conditional transformer language model for controllable
  generation, 2019.
\newblock URL \url{https://arxiv.org/abs/1909.05858}.

\bibitem[Lai et~al.(2017)Lai, Xie, Liu, Yang, and Hovy]{lai-etal-2017-race}
Guokun Lai, Qizhe Xie, Hanxiao Liu, Yiming Yang, and Eduard Hovy.
\newblock {RACE}: Large-scale {R}e{A}ding comprehension dataset from
  examinations.
\newblock In \emph{Proceedings of the 2017 Conference on Empirical Methods in
  Natural Language Processing}, pp.\  785--794, Copenhagen, Denmark, September
  2017. Association for Computational Linguistics.
\newblock \doi{10.18653/v1/D17-1082}.
\newblock URL \url{https://aclanthology.org/D17-1082}.

\bibitem[Lester et~al.(2021)Lester, Al-Rfou, and
  Constant]{lester-etal-2021-power}
Brian Lester, Rami Al-Rfou, and Noah Constant.
\newblock The power of scale for parameter-efficient prompt tuning.
\newblock In \emph{Proceedings of the 2021 Conference on Empirical Methods in
  Natural Language Processing}, pp.\  3045--3059, Online and Punta Cana,
  Dominican Republic, November 2021. Association for Computational Linguistics.
\newblock \doi{10.18653/v1/2021.emnlp-main.243}.
\newblock URL \url{https://aclanthology.org/2021.emnlp-main.243}.

\bibitem[Li et~al.(2021)Li, Nye, and Andreas]{li-etal-2021-implicit}
Belinda~Z. Li, Maxwell Nye, and Jacob Andreas.
\newblock Implicit representations of meaning in neural language models.
\newblock In \emph{Proceedings of the 59th Annual Meeting of the Association
  for Computational Linguistics and the 11th International Joint Conference on
  Natural Language Processing (Volume 1: Long Papers)}, pp.\  1813--1827,
  Online, August 2021. Association for Computational Linguistics.
\newblock \doi{10.18653/v1/2021.acl-long.143}.
\newblock URL \url{https://aclanthology.org/2021.acl-long.143}.

\bibitem[Li \& Liang(2021)Li and Liang]{li-liang-2021-prefix}
Xiang~Lisa Li and Percy Liang.
\newblock Prefix-tuning: Optimizing continuous prompts for generation.
\newblock In \emph{Proceedings of the 59th Annual Meeting of the Association
  for Computational Linguistics and the 11th International Joint Conference on
  Natural Language Processing (Volume 1: Long Papers)}, pp.\  4582--4597,
  Online, August 2021. Association for Computational Linguistics.
\newblock \doi{10.18653/v1/2021.acl-long.353}.
\newblock URL \url{https://aclanthology.org/2021.acl-long.353}.

\bibitem[Loshchilov \& Hutter(2017)Loshchilov and
  Hutter]{Loshchilov2017DecoupledWD}
Ilya Loshchilov and Frank Hutter.
\newblock Decoupled weight decay regularization.
\newblock In \emph{International Conference on Learning Representations}, 2017.

\bibitem[Matthews(1975)]{Matthews1975}
B.W. Matthews.
\newblock Comparison of the predicted and observed secondary structure of t4
  phage lysozyme.
\newblock \emph{Biochimica et Biophysica Acta ({BBA}) - Protein Structure},
  405\penalty0 (2):\penalty0 442--451, October 1975.
\newblock \doi{10.1016/0005-2795(75)90109-9}.
\newblock URL \url{https://doi.org/10.1016/0005-2795(75)90109-9}.

\bibitem[McRae et~al.(2005)McRae, Cree, Seidenberg, and
  Mcnorgan]{mcraeSemanticFeatureProduction2005}
Ken McRae, George~S. Cree, Mark~S. Seidenberg, and Chris Mcnorgan.
\newblock Semantic feature production norms for a large set of living and
  nonliving things.
\newblock \emph{Behavior Research Methods}, 37\penalty0 (4):\penalty0 547--559,
  November 2005.
\newblock ISSN 1554-3528.
\newblock \doi{10.3758/BF03192726}.

\bibitem[Meng et~al.(2022{\natexlab{a}})Meng, Bau, Andonian, and
  Belinkov]{meng2022locating}
Kevin Meng, David Bau, Alex Andonian, and Yonatan Belinkov.
\newblock Locating and editing factual associations in {GPT}.
\newblock \emph{Advances in Neural Information Processing Systems}, 36,
  2022{\natexlab{a}}.

\bibitem[Meng et~al.(2022{\natexlab{b}})Meng, Sen~Sharma, Andonian, Belinkov,
  and Bau]{meng2022memit}
Kevin Meng, Arnab Sen~Sharma, Alex Andonian, Yonatan Belinkov, and David Bau.
\newblock Mass editing memory in a transformer.
\newblock \emph{arXiv preprint arXiv:2210.07229}, 2022{\natexlab{b}}.

\bibitem[Mielke et~al.(2022)Mielke, Szlam, Dinan, and
  Boureau]{mielke2022reducing}
Sabrina~J Mielke, Arthur Szlam, Emily Dinan, and Y-Lan Boureau.
\newblock Reducing conversational agents’ overconfidence through linguistic
  calibration.
\newblock \emph{Transactions of the Association for Computational Linguistics},
  10:\penalty0 857--872, 2022.

\bibitem[Mitchell et~al.(2022)Mitchell, Lin, Bosselut, Finn, and
  Manning]{mitchell2022fast}
Eric Mitchell, Charles Lin, Antoine Bosselut, Chelsea Finn, and Christopher~D
  Manning.
\newblock Fast model editing at scale.
\newblock In \emph{International Conference on Learning Representations}, 2022.
\newblock URL \url{https://openreview.net/pdf?id=0DcZxeWfOPt}.

\bibitem[Paszke et~al.(2019)Paszke, Gross, Massa, Lerer, Bradbury, Chanan,
  Killeen, Lin, Gimelshein, Antiga, et~al.]{paszke2019pytorch}
Adam Paszke, Sam Gross, Francisco Massa, Adam Lerer, James Bradbury, Gregory
  Chanan, Trevor Killeen, Zeming Lin, Natalia Gimelshein, Luca Antiga, et~al.
\newblock Pytorch: An imperative style, high-performance deep learning library.
\newblock \emph{Advances in neural information processing systems}, 32, 2019.

\bibitem[Peng et~al.(2022)Peng, Wang, Hu, Jin, Hou, Li, Liu, and
  Liu]{peng2022copen}
Hao Peng, Xiaozhi Wang, Shengding Hu, Hailong Jin, Lei Hou, Juanzi Li, Zhiyuan
  Liu, and Qun Liu.
\newblock Copen: Probing conceptual knowledge in pre-trained language models.
\newblock In \emph{Proceedings of EMNLP}, 2022.

\bibitem[Perez \& Ribeiro(2022)Perez and Ribeiro]{ignore_previous_prompt}
Fábio Perez and Ian Ribeiro.
\newblock Ignore previous prompt: Attack techniques for language models, 2022.
\newblock URL \url{https://arxiv.org/abs/2211.09527}.

\bibitem[Petroni et~al.(2019)Petroni, Rockt{\"a}schel, Riedel, Lewis, Bakhtin,
  Wu, and Miller]{petroni-etal-2019-language}
Fabio Petroni, Tim Rockt{\"a}schel, Sebastian Riedel, Patrick Lewis, Anton
  Bakhtin, Yuxiang Wu, and Alexander Miller.
\newblock Language models as knowledge bases?
\newblock In \emph{Proceedings of the 2019 Conference on Empirical Methods in
  Natural Language Processing and the 9th International Joint Conference on
  Natural Language Processing (EMNLP-IJCNLP)}, pp.\  2463--2473, Hong Kong,
  China, November 2019. Association for Computational Linguistics.
\newblock \doi{10.18653/v1/D19-1250}.
\newblock URL \url{https://aclanthology.org/D19-1250}.

\bibitem[Radford et~al.(2019)Radford, Wu, Child, Luan, Amodei, and
  Sutskever]{radford2019language}
Alec Radford, Jeff Wu, Rewon Child, David Luan, Dario Amodei, and Ilya
  Sutskever.
\newblock Language models are unsupervised multitask learners.
\newblock 2019.

\bibitem[Rajpurkar et~al.(2016)Rajpurkar, Zhang, Lopyrev, and
  Liang]{rajpurkar-etal-2016-squad}
Pranav Rajpurkar, Jian Zhang, Konstantin Lopyrev, and Percy Liang.
\newblock {SQ}u{AD}: 100,000+ questions for machine comprehension of text.
\newblock In \emph{Proceedings of the 2016 Conference on Empirical Methods in
  Natural Language Processing}, pp.\  2383--2392, Austin, Texas, November 2016.
  Association for Computational Linguistics.
\newblock \doi{10.18653/v1/D16-1264}.
\newblock URL \url{https://aclanthology.org/D16-1264}.

\bibitem[Ravfogel et~al.(2020)Ravfogel, Elazar, Gonen, Twiton, and
  Goldberg]{ravfogel-etal-2020-null}
Shauli Ravfogel, Yanai Elazar, Hila Gonen, Michael Twiton, and Yoav Goldberg.
\newblock Null it out: Guarding protected attributes by iterative nullspace
  projection.
\newblock In \emph{Proceedings of the 58th Annual Meeting of the Association
  for Computational Linguistics}, pp.\  7237--7256, Online, July 2020.
  Association for Computational Linguistics.
\newblock \doi{10.18653/v1/2020.acl-main.647}.
\newblock URL \url{https://aclanthology.org/2020.acl-main.647}.

\bibitem[Ravichander et~al.(2021)Ravichander, Belinkov, and
  Hovy]{ravichander-etal-2021-probing}
Abhilasha Ravichander, Yonatan Belinkov, and Eduard Hovy.
\newblock Probing the probing paradigm: Does probing accuracy entail task
  relevance?
\newblock In \emph{Proceedings of the 16th Conference of the European Chapter
  of the Association for Computational Linguistics: Main Volume}, pp.\
  3363--3377, Online, April 2021. Association for Computational Linguistics.
\newblock \doi{10.18653/v1/2021.eacl-main.295}.
\newblock URL \url{https://aclanthology.org/2021.eacl-main.295}.

\bibitem[Reddy et~al.(2019)Reddy, Chen, and Manning]{reddy-etal-2019-coqa}
Siva Reddy, Danqi Chen, and Christopher~D. Manning.
\newblock {C}o{QA}: A conversational question answering challenge.
\newblock \emph{Transactions of the Association for Computational Linguistics},
  7:\penalty0 249--266, 2019.
\newblock \doi{10.1162/tacl_a_00266}.
\newblock URL \url{https://aclanthology.org/Q19-1016}.

\bibitem[Richardson \& Sabharwal(2020)Richardson and
  Sabharwal]{richardson-sabharwal-2020-qa}
Kyle Richardson and Ashish Sabharwal.
\newblock What does my {QA} model know? devising controlled probes using expert
  knowledge.
\newblock \emph{Transactions of the Association for Computational Linguistics},
  8:\penalty0 572--588, 2020.
\newblock \doi{10.1162/tacl_a_00331}.
\newblock URL \url{https://aclanthology.org/2020.tacl-1.37}.

\bibitem[Ross et~al.(2022)Ross, Wu, Peng, Peters, and
  Gardner]{ross-etal-2022-tailor}
Alexis Ross, Tongshuang Wu, Hao Peng, Matthew Peters, and Matt Gardner.
\newblock Tailor: Generating and perturbing text with semantic controls.
\newblock In \emph{Proceedings of the 60th Annual Meeting of the Association
  for Computational Linguistics (Volume 1: Long Papers)}, pp.\  3194--3213,
  Dublin, Ireland, May 2022. Association for Computational Linguistics.
\newblock \doi{10.18653/v1/2022.acl-long.228}.
\newblock URL \url{https://aclanthology.org/2022.acl-long.228}.

\bibitem[Touvron et~al.(2023)Touvron, Martin, Stone, Albert, Almahairi, Babaei,
  Bashlykov, Batra, Bhargava, Bhosale, Bikel, Blecher, Ferrer, Chen, Cucurull,
  Esiobu, Fernandes, Fu, Fu, Fuller, Gao, Goswami, Goyal, Hartshorn, Hosseini,
  Hou, Inan, Kardas, Kerkez, Khabsa, Kloumann, Korenev, Koura, Lachaux, Lavril,
  Lee, Liskovich, Lu, Mao, Martinet, Mihaylov, Mishra, Molybog, Nie, Poulton,
  Reizenstein, Rungta, Saladi, Schelten, Silva, Smith, Subramanian, Tan, Tang,
  Taylor, Williams, Kuan, Xu, Yan, Zarov, Zhang, Fan, Kambadur, Narang,
  Rodriguez, Stojnic, Edunov, and Scialom]{touvron2023llama}
Hugo Touvron, Louis Martin, Kevin Stone, Peter Albert, Amjad Almahairi, Yasmine
  Babaei, Nikolay Bashlykov, Soumya Batra, Prajjwal Bhargava, Shruti Bhosale,
  Dan Bikel, Lukas Blecher, Cristian~Canton Ferrer, Moya Chen, Guillem
  Cucurull, David Esiobu, Jude Fernandes, Jeremy Fu, Wenyin Fu, Brian Fuller,
  Cynthia Gao, Vedanuj Goswami, Naman Goyal, Anthony Hartshorn, Saghar
  Hosseini, Rui Hou, Hakan Inan, Marcin Kardas, Viktor Kerkez, Madian Khabsa,
  Isabel Kloumann, Artem Korenev, Punit~Singh Koura, Marie-Anne Lachaux,
  Thibaut Lavril, Jenya Lee, Diana Liskovich, Yinghai Lu, Yuning Mao, Xavier
  Martinet, Todor Mihaylov, Pushkar Mishra, Igor Molybog, Yixin Nie, Andrew
  Poulton, Jeremy Reizenstein, Rashi Rungta, Kalyan Saladi, Alan Schelten, Ruan
  Silva, Eric~Michael Smith, Ranjan Subramanian, Xiaoqing~Ellen Tan, Binh Tang,
  Ross Taylor, Adina Williams, Jian~Xiang Kuan, Puxin Xu, Zheng Yan, Iliyan
  Zarov, Yuchen Zhang, Angela Fan, Melanie Kambadur, Sharan Narang, Aurelien
  Rodriguez, Robert Stojnic, Sergey Edunov, and Thomas Scialom.
\newblock Llama 2: Open foundation and fine-tuned chat models, 2023.

\bibitem[Vaswani et~al.(2017)Vaswani, Shazeer, Parmar, Uszkoreit, Jones, Gomez,
  Kaiser, and Polosukhin]{vaswani-transformer}
Ashish Vaswani, Noam Shazeer, Niki Parmar, Jakob Uszkoreit, Llion Jones,
  Aidan~N Gomez, \L~ukasz Kaiser, and Illia Polosukhin.
\newblock Attention is all you need.
\newblock In I.~Guyon, U.~Von Luxburg, S.~Bengio, H.~Wallach, R.~Fergus,
  S.~Vishwanathan, and R.~Garnett (eds.), \emph{Advances in Neural Information
  Processing Systems}, volume~30. Curran Associates, Inc., 2017.
\newblock URL
  \url{https://proceedings.neurips.cc/paper/2017/file/3f5ee243547dee91fbd053c1c4a845aa-Paper.pdf}.

\bibitem[Wang \& Komatsuzaki(2021)Wang and Komatsuzaki]{gpt-j}
Ben Wang and Aran Komatsuzaki.
\newblock {GPT-J-6B: A 6 Billion Parameter Autoregressive Language Model}.
\newblock \url{https://github.com/kingoflolz/mesh-transformer-jax}, May 2021.

\bibitem[Wolf et~al.(2019)Wolf, Debut, Sanh, Chaumond, Delangue, Moi, Cistac,
  Rault, Louf, Funtowicz, et~al.]{wolf2019huggingface}
Thomas Wolf, Lysandre Debut, Victor Sanh, Julien Chaumond, Clement Delangue,
  Anthony Moi, Pierric Cistac, Tim Rault, R{\'e}mi Louf, Morgan Funtowicz,
  et~al.
\newblock Huggingface's transformers: State-of-the-art natural language
  processing.
\newblock \emph{arXiv preprint arXiv:1910.03771}, 2019.

\end{thebibliography}
\bibliographystyle{colm2024_conference}

\appendix

\section{Limitations}
\label{app:limitations}

Our goal in this work has been to demonstrate the expressive power of \ourmethod's representation edits. While we have shown \ourmethod is capable of detecting and mitigating failures in LMs, it has several  limitations that could restrict its usage in production LMs. The foremost is that \ourmethod's linear editing functions must be \emph{learned}, which means users must construct or have access to in-domain training data of the format considered here (each sample has a \emph{prompt}, \emph{entity}, \emph{attribute}, and \emph{target word}). Similarly, using \ourmethod to detect failures of context integration or to detect the absence of prior knowledge requires users to know the correct attribute a priori and to have access to a distractor attribute for comparison; neither may be available in practice. Continued research could expand upon \ourmethod to remove its reliance on training data.

Another limitation of \ourmethod is that the prompting settings considered here, and in all of the closely related \emph{model editing} literature, are deeply simplified for the sake of controlled experimentation. The prompt examples from this paper mostly work out of the box, without \ourmethod, when input into state of the art language models like GPT-4. However, the failure modes we study---factual mistakes and ignoring contextual information---are well documented even in the most performant language models \citep{borji2023categorical}. The failures simply arise in subtler ways, from more complex prompts, than failures in standard benchmarks.

\section{Dataset Preprocessing}
\label{app:datasets}

In \cref{sec:generation}, we evaluate \ourmethod on two dataset (and a third in \cref{sec:mcrae}. Here we detail how they are preprocessed and formatted.

\paragraph{\counterfact} For each record, we use the first paraphrase prompt with the post-edit target object appended to it as the context. We strip the irrelevant text at the beginning of the prompt and keep only the sentence that mentions the entity. We take the attribute to be every token after the entity in the context. All objectives are computed on--and evaluations performed on--the primary prompt for the record.

\paragraph{Bias in Bios} For each record, we take the \emph{second} sentence in the bio longer than three words to be the context.\footnote{The first sentence often explicitly states the person's occupation.} If the sentence does not mention the entity, we prepend the phrase \textit{About [Entity]:} to it. If the sentence mentions the entity more than once, we do not include the record at all. We normalize all mentions of the entity to only use the first name and to not include prefixes like \emph{Dr}. We set the prompt to be \emph{[Entity] has the occupation of}. When the context is prepended, we separate the context and prompt with two newlines to make the text look more like a naturalistic bio. The target word is the person's normalized occupation. Finally, after applying this preprocessing, we randomly sample 5000 records to be in the training set for \ourmethod and 5000 for to be in the held-out evaluation set.

\renewcommand{\feat}{f}
\newcommand{\concept}{c}

\paragraph{McRae Norms} We first compute co-occurrence probabilities for every pair of features in the dataset. For each concept $c$ (e.g., \emph{olive}), the McRae norms data contains a list of features $f_i$ that humans associated with the concept (e.g., \emph{is green}, or \emph{is edible}). The data additionally provides a probability $p(f_i \mid c)$ representing how many people out of thirty ascribed the feature to the concept. Using this data, we sample pairs of features $f_1$ and $f_2$ that co-occur for at least one concept and estimate their co-occurrence probability as follows:
\begin{equation}
\label{eq:mcrae}
   p(\feat_2 \mid \feat_1)
= \frac{p(\feat_1, \feat_2)}{p(\feat_1)}
= \frac{1}{p(\feat_1)} \sum_c p(\feat_2 \mid \concept) p(\feat_1 \mid \concept) p(c) = \frac{1}{N_{f_1}} \sum_c p(\feat_2 \mid \concept) p(\feat_1 \mid \concept) 
\end{equation}
where the sum is over concepts $\concept$, and where $N_{f_1}$ is the number of concepts for which at least one person mentioned $f_1$. Notice that we assume $\feat_1$ and $\feat_2$ are conditionally independent given $\concept$, and that $p(c)$ is uniform. 

We use these human-derived probabilities in two ways. First, when we compute the correlation between $\plm$ and $\phum$, we take $\phum$ to be $p(\feat_1 \mid \feat_2)$ when evaluating against correlated features and $p(\feat_1 \mid c)$ when evaluating against the original features of the concept.
Second, we use $p(\feat_1 \mid \feat_2)$ to filter the set of candidate feature pairs, including only those pairs with co-occurrence probability greater than $.1$.

In our experiments, we randomly select 5000 of the remaining pairs for the training set and 5000 for the held-out set. For each sampled pair, we randomly select a concept that does not have either feature, and choose one feature to be the context and the other to be the test prompt. \ourmethod is trained to maximize the probability of one of the last tokens of the prompt, given the full context as input. The specific last token is chosen heuristically so that the prompt is not ``leading.'' For example, if the prompted feature is \textit{used for eating}, then the target word is \textit{eating}, while if the prompted feature is \textit{grows on trees}, then the target word is \textit{grows}. See the code release for the full implementation.

\section{Training Editors}
\label{app:hyperparameters}

\begin{figure}[t]
    \centering
    \begin{subfigure}{0.49\textwidth}
        \includegraphics[width=\textwidth]{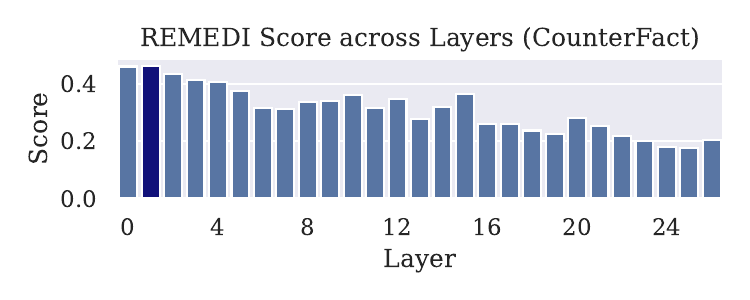}
    \end{subfigure}
    \hfill
    \begin{subfigure}{0.49\textwidth}
        \includegraphics[width=\textwidth]{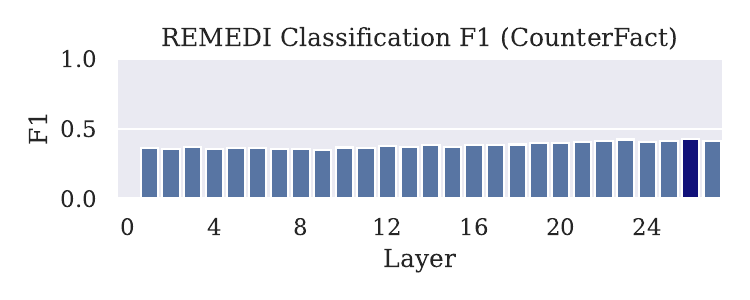}
    \end{subfigure}
    \hfill
    \begin{subfigure}{0.49\textwidth}
        \includegraphics[width=\textwidth]{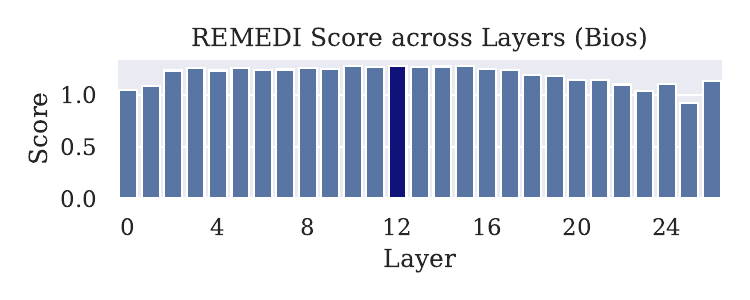}
    \end{subfigure}
    \hfill
    \begin{subfigure}{0.49\textwidth}
        \includegraphics[width=\textwidth]{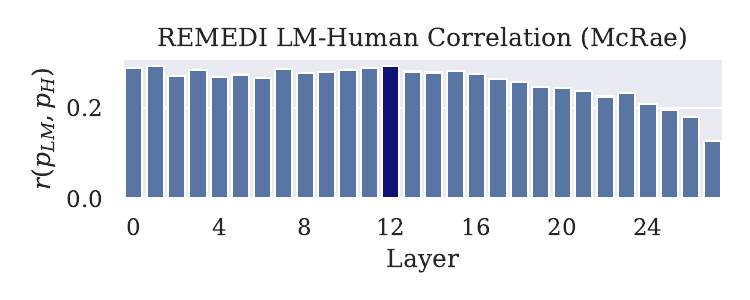}
    \end{subfigure}
    \vspace{-1em}
    \caption{\textbf{Left Column:} Harmonic mean of all the generation quality metrics from \cref{sec:generation} after applying \ourmethod at each layer of GPT-J on a subset of 1000 samples from each dataset. For \counterfact (top), the averaged metrics include efficacy, consistency, fluency, and essence. For Bias in Bios (bottom), it includes accuracy and fluency. \textbf{Upper Right:} \ourmethod classification F1, as described in \cref{sec:probing}, using directions from the best \ourmethod layer for each dataset. In \counterfact, \ourmethod produces the most effective and fluent generations when applied at early layers, while for Bias in Bios it prefers middle layers. \textbf{Upper Right:} classification is most precise when applied to layers after the edit layer. \textbf{Lower Right:} Post-edit human-LM correlation, as defined in \cref{sec:mcrae}, when applying \ourmethod at different layers. \ourmethod works best at earlier layers.}
    \label{fig:layers}
\end{figure}

For both the \counterfact and Bias in Bios datasets, we train $\editor$ using \cref{eq:objective} on a subset of 5000 examples from the dataset, holding out 500 samples for tracking validation loss. For \counterfact, we set $\lamprior = 1$ and $\lamkl = 10$. For Bias in Bios and McRae Norms, we set $\lamprior = 0$ and do not use the $\lossprior$ term. We optimize using AdamW \citep{Loshchilov2017DecoupledWD} with a learning rate of $.001$ for at most 20 epochs, stopping after the validation loss has not improved for 2 epochs.

To decide which layer to apply \ourmethod at, we train editors for every layer in GPT-J and evaluate each on the generation metrics for a subset of 1000 records in the held-out set. For metrics requiring open-ended generation, we use greedy decoding for these sweeps and top-$k$ sampling ($k=5$) in the final evaluations. \cref{fig:layers} (left, bottom) plots the harmonic mean of all generation metrics used in each task (listed in corresponding subsections of \cref{sec:generation}). In \counterfact, earlier layers consistently outperform later layers, suggesting \ourmethod must intervene early to ``override'' knowledge from the LM's pretraining. By contrast, for Bias in Bios and McRae Norms, \ourmethod's performance is relatively flat across early and middle layers. Based on these plots, we chose to apply \ourmethod at layer 1 for \counterfact, and layer 12 for Bias in Bios and McRae.

In \cref{sec:probing}, we measured similarity between \ourmethod directions and entity representations to detect failures in the LM. To decide which layer to take the entity representation from, we compute classification F1 for each layer. Note that the \ourmethod directions fixed to the best layer for generation; we only vary the entity representation layer. Results are shown in \cref{fig:layers} (upper right). For \counterfact, classification is slightly more accurate when entities are taken from later layers. For Bias in Bios, middle layers are best.

\section{Redefining Concepts}
\label{sec:mcrae}

\begin{figure*}[t]
\includegraphics[width=\textwidth]{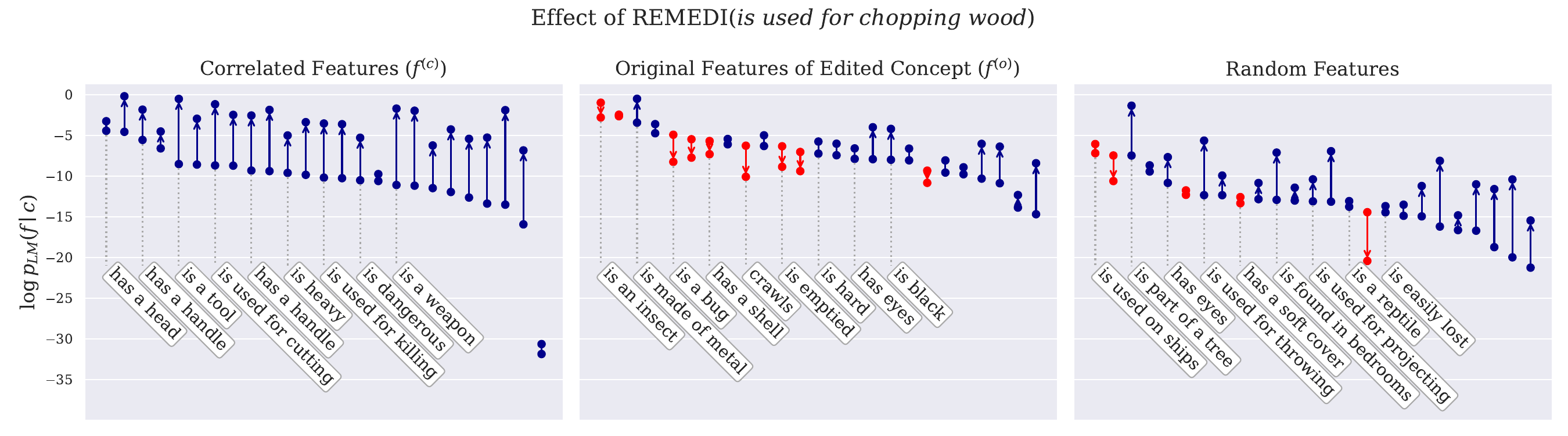}
\vspace{-2em}
\caption{Change in LM log-probability for different feature strings after using \ourmethod to add the feature \textit{is used for chopping wood} to seven different concepts. Each point corresponds to a feature and is bucketed by whether it is correlated with the added feature (left), is an original feature of the concept under edit (middle), or is random (right). Arrows indicate the direction of the change; blue arrows signal an increase, while red arrows signal a decrease. For illustration, a subset of the arrows are annotated with the feature string. 
  }
\label{fig:mcrae}
\end{figure*}

In our final set of generation experiments, we use \ourmethod to edit basic noun concepts (like \emph{olive} or \emph{airplane}) and change their definitions.
Noun concepts are typically defined by the set of \emph{features} language users associate with them: olives can be green and often appear in salads; airplanes are largely made of metal and can fly; and spiders have eight legs and spin webs. 

Our experiments use \ourmethod to \emph{add features} to concepts, then study the effect of these concept modifications on other related features. 
We use common nouns (\textit{olive}) as edit targets, and feature descriptions (\textit{is made of metal}) as attributes. 
Properties like \emph{is made of metal}, \emph{is hard}, and \emph{is shiny} exist in a complex network of entailment and correlation relations, and we are interested in characterizing whether \ourmethod respects these associations (e.g.\ increasing the probability of the string \emph{olives are inedible} after increasing the probability of the string \emph{olives are made of metal}).

\newcommand{\conc}{c}
\newcommand{\featorig}{f^{(o)}}
\newcommand{\featcorr}{f^{(c)}}
\newcommand{\featnew}{f^*}

\paragraph{Setup}
We obtain concepts and features from the McRae Norms dataset \citep{mcraeSemanticFeatureProduction2005}. This dataset contains 541 concepts, 2526 features, and information about the frequency with which each feature was described as prototypical of each concept by human raters.
We construct a dataset containing 10k entries, split evenly into train and test sets, where each entry consists of a concept $\conc$, a list of \emph{original} features $\featorig$ for the concept, a target feature to add $\featnew$, and a list of features $\featcorr$ that are \emph{correlated} with the new feature. Details about data and hyperparameters are in Appendices~\ref{app:datasets} and \ref{app:hyperparameters}.

\paragraph{Metrics}
We measure average absolute change in probability for correlated and original features. If $\feat$ is any held out feature string ($\featorig$ or $\featcorr$), we define absolute change as:
\begin{equation}
    \label{eq:rel-change}
    \Delta \plm(\feat \mid \conc, \featnew) = \lm(\feat \mid \conc, \featnew) -
    \lm(\feat \mid \conc) ~ ,
\end{equation}
where $\lm(\cdot)$ denotes the probability that the LM assigns to $f$ conditioned on $c$ as a prompt and with $\featnew$ added to the concept via textual prompting or via \ourmethod. We additionally measure the correlation between LM probabilities and human-derived probabilities $\phum(\feat)$ for held-out features, which we denote $r(\plm, \phum)$. For \emph{original} features, we compute $\phum(\featorig)$ as the proportion of human annotators who described $\featorig$ as a prototypical feature of the concept being edited. For \emph{correlated} features, we compute $\phum(\featcorr)$ as the co-occurrence probability with the feature being inserted.

\begin{table}[t]
    \centering
    \footnotesize
    \begin{tabular}{lccccc}
        \toprule
        & \multicolumn{2}{c}{\textbf{Correlated}}
        & \multicolumn{2}{c}{\textbf{Original}}
        & \textbf{Rand.}   \\
        \midrule
        Method &
        $\Delta \plm$
        & $r$
        & $\Delta \plm$
        & $r$
        & $\Delta \plm$ \\
        \midrule
        No Edit & -- & $.11$ & -- & $.26$ & -- \\
        Prefix & $0.4$ $(0.7)$ & $.16$ & $0.0$ $(1.7)$ & $.25$ & 0.0 (0.0) \\
        \ourmethod & $7.1$ $(5.2)$ & $.29$ & $0.5$ $(3.6)$ & $.19$ & $0.2$ $(0.9)$ \\
        \bottomrule
    \end{tabular}
    \caption{
        Comparison between \ourmethod and a prefix baseline for adding new
features to concepts from 
\citet{mcraeSemanticFeatureProduction2005}. $\Delta \plm$ is the mean (SD) of
  the absolute change in LM probability assigned to feature strings, scaled by
  100. $r$ is shorthand for $\corr$, the correlation between the
  post-intervention LM probabilities for features and their human-derived
  counterparts. Compared to prefixing, \ourmethod causes a large increase in
  $\plm$ for all correlated features, as well as modest changes to original
  features in either direction. On random, unrelated features, both methods have
  little effect. \ourmethod nearly triples the LM's correlation with human
  feature relatedness judgments.
    }
    \label{tab:mcrae-correlations}
\end{table}

\paragraph{Results}
\Cref{tab:mcrae-correlations} compares \ourmethod to the prefix baseline where
the new attribute (e.g.\ \textit{An olive is made of metal}) is prepended to the prompt. Using \ourmethod results in a much stronger effect than prefixing: correlated features see an order of magnitude larger increase in probability and become substantially more correlated with the human-derived feature co-occurrence probabilities. This suggests that \ourmethod preserves the correlates of added attributes: an \emph{olive}, now \emph{made of metal}, is more likely to be \emph{shiny}.

\ourmethod has a slightly subtler impact on the concept's original features. The near-zero mean and large standard deviation highlight that some original features are promoted under \ourmethod while others are suppressed, likely because they conflict with the added feature (e.g.\ \emph{olives} cannot be both \emph{made of metal} and \emph{edible}). This is further reflected in the decrease of $\corr$: the language model's post-edit distribution over a concept's original features less resembles the human distribution.
Finally, \ourmethod has a negligible effect on the probabilities assigned to random, unrelated features, indicating that the edits primarily impact the relevant feature associations.

\Cref{fig:mcrae} provides a concrete example of \ourmethod's effect on $\plm$ when adding the \textit{is used for chopping wood} feature. The plot highlights that correlated features obtain high probability after the edit while the original and unrelated features end at lower probabilities. Taken together, these results demonstrate that \ourmethod edits can be applied not only to named entities but also to generic noun concepts, and these edits modify concepts' relations globally rather than simply priming the LM to produce specific target text.

\section{Results in Other Models}
\label{app:other_models}

We run the evaluations of \ourmethod from Sections~\ref{sec:generation} and \ref{sec:probing} on two additional models: GPT2-XL \citep{radford2019language}, which has fewer parameters than GPT-J (1B vs. 7B) and Llama-2-13b \citep{touvron2023llama}, which has more (13B vs. 7B). We select layers to perform editing and classification layers using the same procedures as before. Results are in Tables~\ref{tab:generation_bias_other_models}, \ref{tab:generation_facts_other_models}, and \ref{tab:probing_other_models}.

\begin{table}[ht]
    \centering
    \begin{tabular}{llcccc}
        \toprule
        \textbf{Model} & \textbf{Method} & \textbf{Ctx Acc.} & \textbf{Ctx Flu.} & \textbf{No-Ctx Acc.} & \textbf{No-Ctx Flu.} \\
        \midrule
        GPT2-XL & LM only & 0.079 & 560.1 & 0.01 & 646.6 \\
        & \ourmethod & 0.084 & 539.8 & 0.27 & 554.7 \\
        \midrule
        Llama-2-13b & LM only & .71 & 491.3 & .04 & 625.7 \\
        & \ourmethod & .72 & 457.1 & .64 & 599.5 \\
        \bottomrule
    \end{tabular}
    \caption{Results on the error correction task of \cref{sec:generationerrors} for two different models. Mirrors \cref{tab:errorcorrection}.}
    \label{tab:generation_bias_other_models}
\end{table}

\begin{table}[ht]
\centering
\begin{tabular}{@{}llccccc@{}}
\toprule
\textbf{Model}        & \textbf{Approach} & \textbf{Eff}  & \textbf{Nbr}  & \textbf{Cons} & \textbf{Flu}   & \textbf{Ess}  \\ \midrule
GPT2-XL      & Prefix   & .84  & 1.0  & .24  & 594.1 & .39  \\
             & Replace  & .75  & 1.0  & .33  & 609.3 & .08  \\
             & \ourmethod   & .97  & 1.0  & .32  & 597.0 & .28  \\
             \midrule
Llama-2-13b  & Prefix   & .84  & 1.0  & .17  & 557.0 & .41  \\
             & Replace  & .86  & 1.0  & .30  & 591.8 & .06  \\
             & \ourmethod   & .98  & 1.0  & .25  & 550.5 & .29  \\ \bottomrule
\end{tabular}
\caption{Results on the factual editing task of \cref{sec:generationfacts} for two different models. Mirrors \cref{tab:counterfact-edit}.}
\label{tab:generation_facts_other_models}
\end{table}

\begin{table}[ht]
\centering
\begin{tabular}{@{}lccccc@{}}
\toprule
\textbf{Model}            & \textbf{Approach}    & \textbf{Ctx F1} & \textbf{Ctx MCC} & \textbf{No-Ctx F1} & \textbf{No-Ctx MCC} \\ \midrule
GPT2-XL          & Identity    & .37    & .05     & .46       & .11        \\
                 & Fact probe  & .32    & .00     & .45       & .00        \\
                 & Shortcut    & .34    & .05     & .45       & -.02       \\
                 & \ourmethod      & .33    & .01     & .48       & .05        \\ \midrule
Llama-2-13b & Identity   & .25    & -.01    & .17       & .00        \\
                 & Fact probe  & .23    & -.03    & .17       & .02        \\
                 & Shortcut    & .24    & -.00    & .16       & -.01       \\
                 & \ourmethod      & .29    & .07     & .28       & .19        \\ \bottomrule
\end{tabular}
\caption{Results on the error detection task of \cref{sec:probing} for two different models. Mirrors \cref{tab:classification}.}
\label{tab:probing_other_models}
\end{table}

In general, the same trends observed in GPT-J hold across GPT2-XL and Llama-2-13b, with greater effectiveness as model size increases. For generation tasks, \ourmethod exerts more effective control than standard prompting. For error classification tasks, \ourmethod transfers effectively from non-contextual training to contextual classification in all models.

\cref{tab:generation_bias_other_models} reveals several qualitative differences between models. In particular, GPT2-XL is extremely ineffective at following the prompt on the BiasInBios task (\cref{tab:generation_bias_other_models} Ctx Acc.), while \ourmethod, applied without a textual context, is substantially better at steering model behavior in such a small model. \ourmethod is less effective when a textual context is also provided, but still improves slightly upon the baseline. By contrast, Llama-2-13b consistently follows the prompt at baseline, and exhibits a relatively low error rate to begin with. \ourmethod, applied in context, improves accuracy by a small amount, but by much less than when the baseline LM is ineffective.

\section{Analyzing \ourmethod Edits}
\label{app:analysis}

\subsection{Failure Modes}

\begingroup
\renewcommand{\arraystretch}{1.5} %
\begin{table*}[t]
    \centering
    \scriptsize
    \begin{tabular}{p{2.75cm} p{4cm} p{5.75cm}}
        \toprule
        \textbf{Failure mode (Generic)} & \textbf{Context} & \textbf{Generation} \\
        \midrule
        Repeats indefinitely & About Joy: She is also a program faculty member of CWRU’s Women’s and Gender Studies and Ethnic Studies programs. & Joy has the occupation of professor of psychology, professor of sociology, professor of anthropology, professor of English, professor of communication, professor of journalism, professor of music\dots \\
        Incorrect edit & About Mason: He is a graduate of Marquette University School of Dentistry. & Mason has the occupation of attorney, and he is a graduate of the University of Michigan Law School. \\
        Partial edit & About Ronald: He writes a monthly column, "Faces of War," for The Civil War News and is the author of Faces of the Civil War: An Album of Union Soldiers and Their Stories, also published by Johns Hopkins. & Ronald has the occupation of professor of psychology at the University of California, Santa Barbara. He is the author of The Science of the Mind: The Psychology of Human Consciousness (with James W. Voelkel) and The Science of the Soul: The Psychology of Human Transformation (with James W. Voelkel). \\
        \midrule
        \textbf{Failure Mode (Famous)} & & \\
        \midrule
        Repeats indefinitely & Which position does Viktor Klonaridis play? They play as quarterback & Viktor Klonaridis's greatest strength is quarterback quarterback quarterback quarterback quarterback \\
        Destroys essence & Wiener schnitzel, from Japan & Wiener schnitzel is a Japanese anime series that was broadcast in Japan. It was produced by Japan Animation Association (J.A.P.) and was broadcast in Japan. \\
        Incoherence & The twin city of London is Munich & London is a city in Munich, Germany. It is the capital of Munich and the second largest city in Germany. \\
        Changes unrelated facts & My Name Is Earl was released on CBS & My Name Is Earl is a CBS television series that aired from September 1972 to May 1973. It was the first television series to be broadcast in color. \\
        \bottomrule
    \end{tabular}
    \caption{Examples of \ourmethod's failure modes in Bias in Bios (top) and \counterfact (bottom). In both settings, \ourmethod occasionally causes disfluent or incoherent generations where the model to repeats itself indefinitely. On generic entities, \ourmethod sometimes (though rarely) will make an incorrect edit (e.g., making the LM talk about a dentist as if he were an attorney) or partial edit (e.g., correctly editing in that \textit{Ronald} is a professor, but missing that he is a professor of \emph{history}). On famous entities, \ourmethod can sometimes damage the essence of the entity (e.g., by making \textit{Wiener schnitzel} an anime instead of a food), cause further incoherence (e.g., by making \textit{Munich} cities have sub-cities), or accidentally change related facts (e.g., by changing the air dates of \textit{My Name is Earl}).}
    \label{tab:failure-modes}
\end{table*}
\endgroup

\Cref{tab:failure-modes} shows examples of \ourmethod's failure modes, taken from the evaluations of \Cref{sec:generationfacts}. While Tables \ref{tab:errorcorrection} and \ref{tab:counterfact-edit} show that \ourmethod is effective at causing the LM to generate text consistent with the attribute, the act of editing the LM's representations can occasionally lead to disfluent or incorrect generations. In generic entities, these cases primarily involve \ourmethod failing to insert the attribute, or only inserting a part of it. In famous entities, \ourmethod sometimes damages the essence of the entity, leading the LM to generate text that is consistent with the new attribute but not consistent with any \emph{original} attribute of the entity, as in the \textit{Wiener schitzel} and \textit{Munich} examples. \ourmethod can also cause unrelated facts to change, such as the airtime of \textit{My Name is Earl} in the bottom row.

Some of these errors might originate from the model itself. We observe disfluent, repeating generations even when we do not apply \ourmethod and only prepend the context to its input. Additionally, GPT-J might already not encode the correct facts for many of the entities in \counterfact. Nevertheless, these errors could potentially be mitigated by training \ourmethod's editing models on larger datasets or by editing at different or multiple layers.

\subsection{Generalization to Unseen Attributes}

During the \counterfact evaluation from \Cref{sec:generationfacts}, we test \ourmethod on held out (\emph{entity}, \emph{attribute}) pairs. However, we can also consider how well \ourmethod generalizes to just new attributes, regardless of which entity they were edited into.

\begin{table}[ht]
    \centering
    \begin{tabular}{lccccc}
        \toprule
        \textbf{Setting} & Total & Efficacy $\uparrow$ & Consistency $\uparrow$ & Fluency $\uparrow$ & Essence $\uparrow$ \\
        \midrule
        Seen in Training & $3311$ & $99.5$ & $29.5$ & $474.5$ & $23.9$ \\
        Unseen in Training & $1689$ & $95.5$ & $30.8$ & $508.6$ & $26.5$ \\
        \midrule
        Model Knows & $4184$ & $98.0$ & $30.5$ & $486.2$ & $25.2$ \\
        Model Does Not Know & $816$ & $98.8$ & $27.3$ & $485.0$ & $22.5$ \\
        \bottomrule
    \end{tabular}
    \caption{\ourmethod editing metrics on \counterfact, broken down by whether the attribute appeared in \ourmethod's training data (top) and whether the GPT-J correctly predicts the true fact given the prompt without any intervention (bottom). While \ourmethod is slightly less effective at overwriting the original fact with unseen attributes, it still produces a correct edit over 95\% of the time and even causes substantially more fluent and essence-preserving generations in this setting. \ourmethod is also slightly more effective at editing entities for which the LM has a strong prior, though the subsets are relatively unbalanced and this could be due to noise.}
    \label{tab:cf-breakdown}
\end{table}

The top half of \Cref{tab:cf-breakdown} shows \ourmethod's performance on the \counterfact benchmark broken down by whether the target attribute was seen during training, as determined by exact string match. While slightly less efficacious, \ourmethod performs best on all other metrics when the attribute was not seen during training. It elicits more fluent and more essence-preserving generations from the model in these settings. This difference could arise from overfitting of the linear editor.

\subsection{Effect of Prior Knowledge}

Additionally, when using \ourmethod to edit factual knowledge, we can ask how sensitive it is to whether the language model encodes the correct fact prior to editing. The bottom half of \Cref{tab:cf-breakdown} shows performance on \counterfact broken down by whether the language model correctly ranks the true object for the fact (\textit{Paris} in the prompt \textit{The Eiffel Tower is located in}) ahead of a distractor object (\textit{Rome}). We see that \ourmethod performs slightly better when the language model does know the correct entity. Specifically, in these settings, \ourmethod is better at preserving the entity's essence, like because the language model has a very strong opinion about what the entity is.

\subsection{\ourmethod Direction Norms}

\begin{figure}[t]
    \centering
    \begin{subfigure}{0.49\textwidth}
        \includegraphics[width=\textwidth]{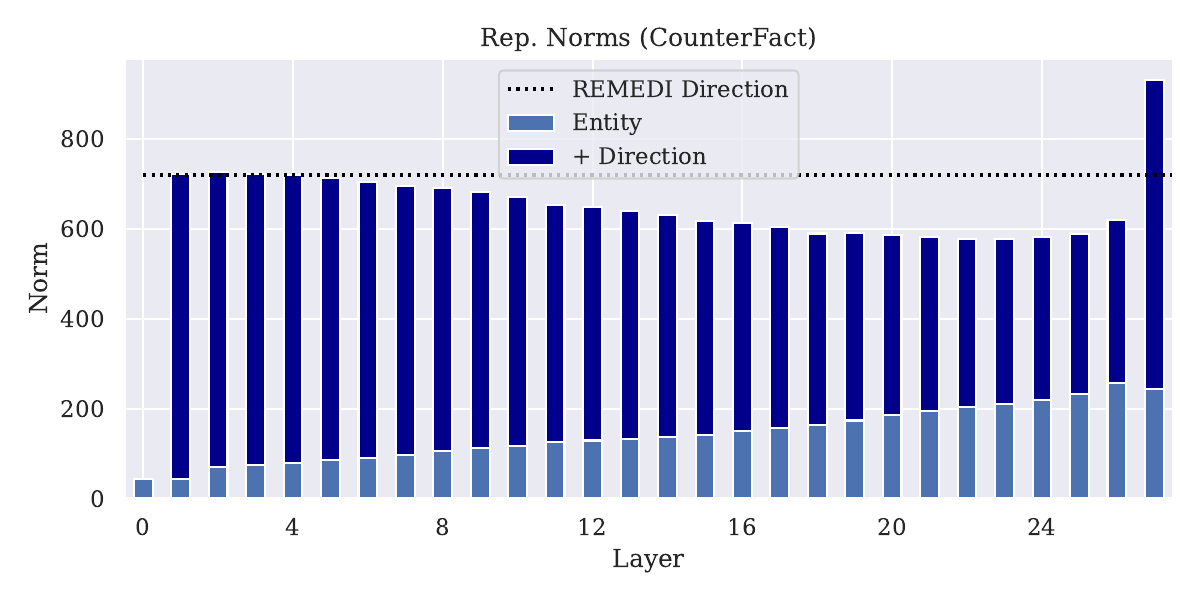}
    \end{subfigure}
    \begin{subfigure}{0.49\textwidth}
        \includegraphics[width=\textwidth]{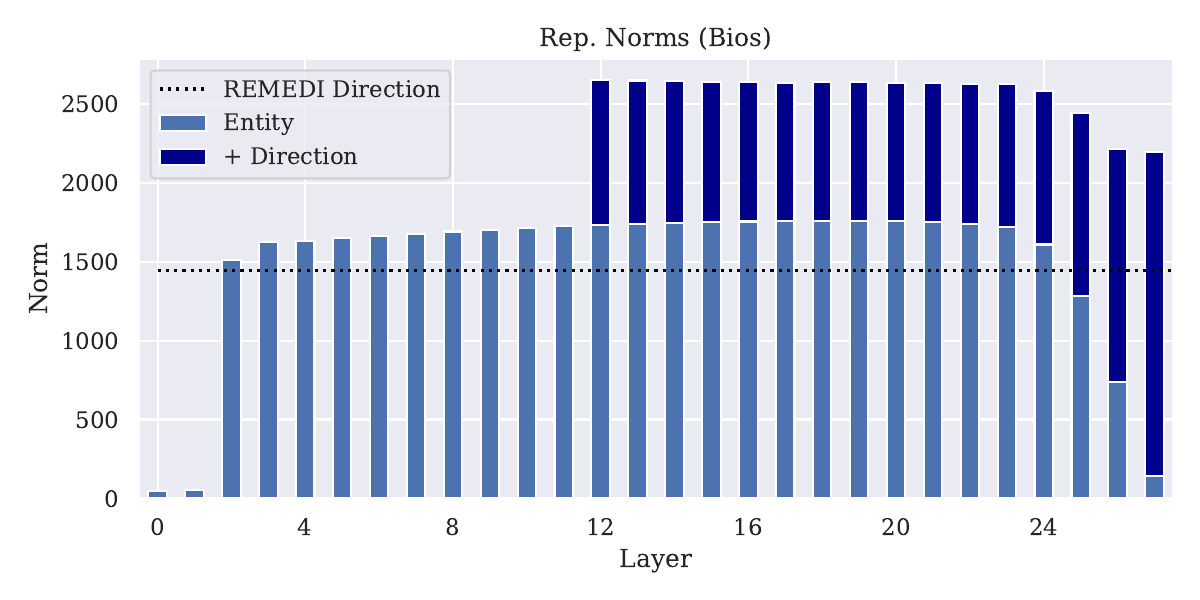}
    \end{subfigure}
    \begin{subfigure}{0.49\textwidth}
        \includegraphics[width=\textwidth]{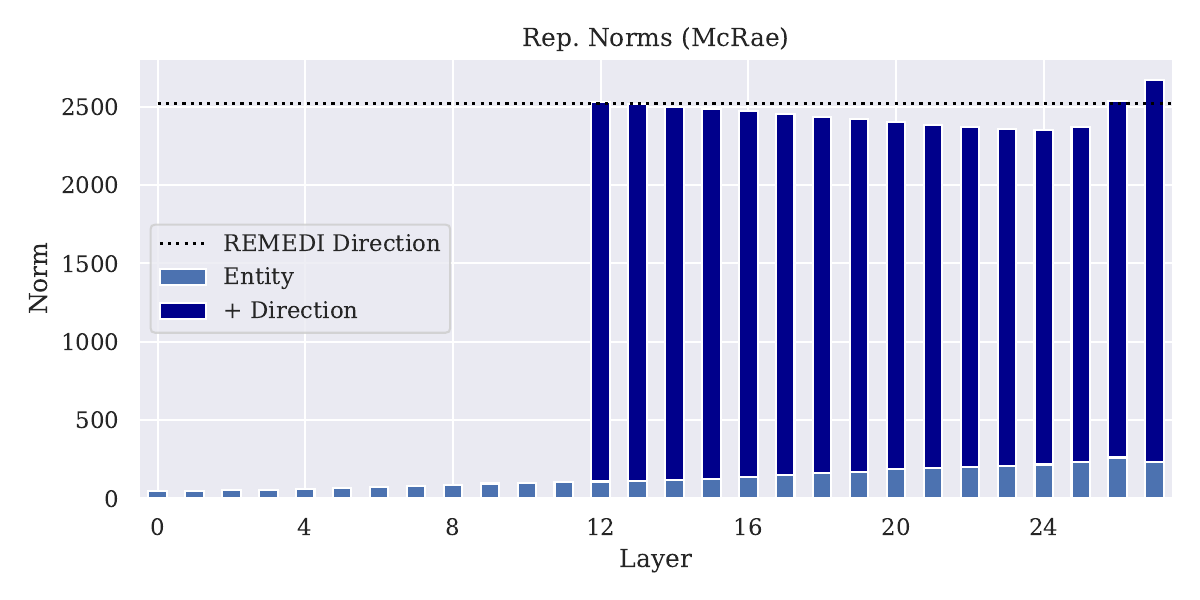}
    \end{subfigure}
    \caption{Average representation norm of the entity representation across GPT-J layers before (light blue) and after (dark blue) editing at the optimal layer. In the factual editing and concept editing settings, the \ourmethod edit direction is many times larger than the entity's representation, while for the non-famous entities of Bias in Bios the average direction is much smaller.}
    \label{fig:norms}
\end{figure}

Recall that \ourmethod involves adding a direction, which captures the target attribute, to an LM's representation of an entity. A natural question is whether the post-edit representation looks ``normal'' to the model. We observe that the norms of \ourmethod directions are quite large relative to the model's hidden states at the layer being edited. This is illustrated in \cref{fig:norms}. When applying \ourmethod to \counterfact and McRae Norms samples, the directions are substantially larger than the edit target's representations, and consequently the edited representation is sometimes more than twice as large as it was pre-edit. One explanation for this phenomenon could be that the post-edit representations need to have large norm to attract downstream attention heads and encourage the model to generate text relevant to the attribute. Indeed, \ourmethod's objective (see \cref{eq:objective}) explicitly rewards the model for not just encoding the target attribute, but for making the LM generate text about it. However, it is not clear that \ourmethod directions or the edited representations are \emph{abnormally} large to the model. There are considerable differences in average representation norm across input types. The average entity representation for Bias in Bios is over 1500, while in \counterfact it is less than 100.

\section{Full Prompts for Qualitative Examples}

\begingroup
\renewcommand{\arraystretch}{1.5}
\begin{table}[t]
    \centering
    \scriptsize
    \begin{tabular}{p{1cm} p{3cm} p{3.75cm} p{3.75cm}}
        \toprule
         \textbf{Entity} & \textbf{Context} & \textbf{No Edit} & \textbf{\ourmethod}  \\
         \midrule
         Anita & Anita's legal office serves the lower Eastern Shore including Accomack and Northampton counties. & [Context]\verb=\=n\verb=\=nAnita has the occupation of a Licensed Practical Nurse. She has been practicing law for over 30 years. & Anita has the occupation of attorney. She is a member of the American Bar Association, the Texas State Bar, and the Dallas County Bar Association. \\
         London Bridge &
         The London Bridge is located in the deserts of Arizona. &
         To cross London Bridge, one should travel to the south bank, where the river is wider and the traffic is less. & To cross London Bridge, one should travel to Arizona. \\
         Gianni Versace S.p.A. & Gianni Versace S.p.A.'s headquarters is surrounded by London. & [Context] The headquarters of Gianni Versace S.p.A. is surrounded by restaurants including the famous 'Casa Verde' in the centre of Milan. & The headquarters of Gianni Versace S.p.A. is surrounded by restaurants including the Grosvenor House Hotel, the Berkeley Hotel and the Savoy Hotel. \\
         \bottomrule
    \end{tabular}
    \caption{Full prompts and GPT-J outputs for the examples shown in \Cref{fig:teaser}. Note that the \textit{Anita} and \textit{Versace} examples include the context in the prompt to illustrate failures of context integration, while the \textit{London Bridge} example does not in order to illustrate how GPT-J encodes prior knowledge about famous entities.}
    \label{tab:full-prompts}
\end{table}
\endgroup

\Cref{fig:teaser} includes several qualitative examples which are shortened for space and exposition. The full prompts and GPT-J outputs, before and after applying \ourmethod, are shown in \Cref{tab:full-prompts}. All qualitative examples in this paper (\Cref{fig:teaser} and Tables~\ref{tab:qualitative}, \ref{tab:failure-modes}, and \ref{tab:full-prompts}) were generated using greedy decoding.

\end{document}